\documentclass[default]{article}

\usepackage{geometry}
\usepackage{graphicx}%
\usepackage{multirow}%
\usepackage{amsmath,amssymb,amsfonts}%
\usepackage{amsthm}%
\usepackage{mathrsfs}%
\usepackage[title]{appendix}%
\usepackage{xcolor}%
\usepackage{textcomp}%
\usepackage{manyfoot}%
\usepackage{booktabs}%
\usepackage{algorithm}%
\usepackage{algorithmicx}%
\usepackage{algpseudocode}%
\usepackage{listings}%
\usepackage{setspace}
\usepackage[colorinlistoftodos]{todonotes}
\usepackage[font=small,labelfont=bf]{caption}
\usepackage{enumerate}
\usepackage[autostyle]{csquotes}
\usepackage[table]{colortbl}
\usepackage{ulem}
\usepackage{hhline}
\usepackage{placeins}
\usepackage{float}
\usepackage{subcaption}
\usepackage{hyperref}

\DeclareMathOperator*{\argmin}{arg\,min}
\DeclareMathOperator{\sign}{sign}

\newtheorem{theorem}{Theorem}
\newtheorem{lemma}{Lemma}

\newcommand{\R}{\mathbb{R}}
\newcommand{\N}{\mathbb{N}}
\newcommand{\Z}{\mathbb{Z}}
\providecommand{\C}{\mathbb{C}}

\begin{document}
\title{Riesz feature representation: scale equivariant scattering network for classification tasks} 

\author{Tin Barisin\thanks{RPTU Kaiserslautern-Landau, Kaiserslautern, Germany.}
\and Jesus Angulo\thanks{Center for Mathematical Morphology
MINES Paris, PSL Research University, Fontainebleau,
France.}
\and Katja Schladitz\thanks{Fraunhofer-Institut für Techno- und Wirtschaftsmathematik (ITWM), Kaiserslautern, Germany.}
\and Claudia Redenbach\footnotemark[1]}

\maketitle

\begin{abstract}
    Scattering networks yield powerful and robust hierarchical image descriptors which do not require lengthy training and which work well with very few training data. 
    However, they rely on sampling the scale dimension. Hence, they become sensitive to scale variations and are unable to generalize to unseen scales. 
    In this work, we define an alternative feature representation based on the Riesz transform. We detail and analyze the mathematical foundations behind this representation. In particular, it inherits scale equivariance from the Riesz transform and completely avoids sampling of the scale dimension. Additionally, the number of features in the representation is reduced by a factor four compared to scattering networks.
    Nevertheless, our representation performs comparably well for texture classification with an interesting addition: scale equivariance. Our method yields 
    very good
    performance when dealing with scales outside of those covered by the training dataset. The usefulness of the equivariance property is demonstrated on the digit classification task, where accuracy remains stable even for scales four times larger than the one chosen for training. As a second example, we consider classification of textures. Finally, we show how this representation can be used to build hybrid deep learning methods that are more stable to scale variations than standard deep networks.
\end{abstract}

This work was funded by the German Federal Ministry of Education and Research (BMBF) [grant number 05M2020 (DAnoBi)].



\section{Introduction}
Deep learning offers very powerful tools for standard tasks in computer vision and image processing such as image segmentation or classification. However, one of its main drawbacks is that it is data-hungry, i.e. it requires a large amount of annotated data with enough inter-class variability for the training part to be successful. Data collection and processing are time-consuming and in some cases it is not easy to gather a sufficient amount of suitable data. For cases where very few data is available, feature extraction combined with a classifier like principal component analysis (PCA) or a support vector machine (SVM) becomes more practical as well as more efficient in terms of computation and energy consumption.

Hierarchical feature representations are an efficient way to 
extract universal features that can be used for a wide range of tasks. 
The idea originates from so-called scattering networks~\cite{mallat2012}, where a complex wavelet 
is used to extract the local amplitude. Rotated and rescaled versions of this wavelet capture diverse features and serve as one layer of the representation.
The hierarchical representation then arises from applying the same set of wavelets on the amplitudes from the previous layer. In this way, higher order nonlinear features can be extracted and used as input to a classifier. 

Generally, it is desirable that any feature representation satisfies basic properties of human vision and is able to handle acquisition variability of 2d digital cameras. That is, we require translation (shift) 
and scale (dilation) invariance\ as well as local rotation invariance.
These properties express that feature representations should not depend on objects' position, 
size in the image, or small changes in the angle of view, i.e. they should be robust to variations in these geometric features.
However, achieving these invariances is not trivial at all. 
Here, we impose translation invariance as a basic condition that a representation should satisfy and investigate how to additionally achieve robustness to scale changes or even scale invariance.

Mathematically, scale-space is a continuous semi-group, so scales are unlimited.
However, 
images are discretized and bounded representations of the real world and for that reason scale in images is bounded from two sides~\cite{lindeberg2020}. From below, the bound is introduced by image resolution or pixel size (inner scale). Details smaller than the pixel size cannot be recorded. From above, image scale is bounded by the image size (outer scale). What happens outside of the image cannot be observed. 
A typical approach for obtaining scale invariance is to sample the scale dimension and to extract features at each sampled scale, e.g. in the SIFT descriptor~\cite{lowe1999object}.
However, different scales in the image are not independent and many of them are not relevant for the given task such that
a uniform sampling of the scale dimension results in highly redundant representations. 
Another problem arises from scale sampling when the window size is changed. In that case, we might need more or less scales for the image, depending on whether the window size is decreased or increased. Furthermore, when the pixel size changes, scale sampling has to be adjusted to derive a comparable representation.
For these reasons, scale sampling may require adaptations and bears potential risks 
during automatic feature extraction.

As a non-local, scale equivariant operator, the Riesz transform \cite{barisin22riesz} can serve as a replacement to scale sampling.
Scale equivariance means that the Riesz transform commutes with the scaling operator.
As a consequence, all scales can be accounted for simultaneously using a single operator.
Another benefit of the Riesz transform is that it automatically adjusts to the image size. Hence, it represents a 
simple way to analyze multiscale data without sampling of the scale dimension. 

In this work, we design a hierarchical feature representation based on the Riesz transform as an alternative to scattering networks. The Riesz transform can be used to construct a quadrature filter for signal decomposition into amplitude and phase. Computing the amplitude acts as a nonlinear transformation and is used to build layers in depth. Our representation inherits the scale equivariance property of the Riesz transform. Global average pooling at the end results in a scale and translation invariant hierarchical representation. 
Theoretical implications of this representation are discussed and its application for texture and digit classification is presented. A hybrid deep learning method based on this representation is built and used for object classification in natural images. In these applications, robustness to scale variations is observed. Furthermore, our representation generalizes to completely unseen scales that are not covered by the training set.

\subsection{Equivariance and invariance}
We now formally define and intuitively interpret equivariance and invariance.
Let $\mathscr{L}=\{f:\R^d\to\R\}$ be a set of functions that we refer to as images. Let $(G,.)$ be a group or semi-group. We consider a family of operators $L_g: \mathscr{L} \to \mathscr{L}$ that transform an image $ f \in \mathscr{L}$ based on a group element $g\in G$ via $L_g(f) = f(x.g^{-1})$. 
The operator $F: \mathscr{L} \to \mathscr{L}$ is said to be \emph{equivariant} with respect to $L_g$ if
$$ F(L_g(f)) = L_g(F(f)) \text{ for all } f \in \mathscr{L}$$
and \emph{invariant} to $L_g$ if
$$ F(L_g(f)) = F(f) \text{ for all } f \in \mathscr{L}.$$

Invariance is a stronger property than equivariance. It implies that applying operator $L_g$ does not have any effect on the result of operator $F$.
In the context of images, equivariance implies that the order in which we apply the operators $F$ and $L_g$ on the input image does not matter. Invariance implies that the operator $F$ disregards
the effect of operator $L_g$.
In applications, it is useful to have an equivariant feature extractor as an input to a classifier. Classification should then be invariant to certain transformations, i.e. the class label should not be changed when applying these transformations. In image processing, we usually want to achieve invariance to translation, rotation, and/or rescaling.

\section{Related work}

\subsection{Scattering networks and related work}
The idea behind scattering networks is to design hierarchical image descriptors similar to CNNs which are robust or (locally) invariant to transformations from a fixed compact Lie group~\cite{mallat2012}. Details on scattering networks can be found in Appendix~\ref{appendix:scattering}. 

The design is based on constructing a nearly quadrature filter from Morlet or other types of wavelets, i.e., a filter that decomposes the signal into amplitude and phase.
Since the estimation of these characteristics is intrinsically noisy, various quadrature filters have been devised, see~\cite{boukerroui04} for an overview.  The filtering is repeated for several rescaled and rotated versions of the Morlet wavelet. Only amplitude information is kept, and the same set of wavelets is applied to the amplitudes sequentially to derive higher order features.

An important result of Mallat~\cite{mallat2012} is that the scattering network is Lipschitz continuous with respect to the actions from the diffeomorphism group. This means that the scattering network yields very similar outputs for the original image and "slightly locally" deformed versions of it.
As a first application example, a scattering network on the translation group was used to construct a (locally) translation invariant feature representation which was applied to texture and digit classification problems~\cite{bruna2011,bruna2012-2}. This representation was further extended to local invariance for translations, rotations, and scalings~\cite{sifre2013}, which proves to be useful for unknown viewing conditions in texture classification. 
Furthermore, a scattering network that is stable to rotations and deformations along the rotation axes was applied to solve object classification tasks in~\cite{oyallon2015}.

While scattering networks have shown to perform well on datasets with little training data (e.g. with 46 images per class on the CUReT dataset in~\cite{bruna2012-2}), this class of methods is still outperformed by CNNs on large datasets such as Caltech or CIFAR ~\cite{oyallon2015}.
A possible explanation for this has been suggested in~\cite{cotter2017visualizing}: scattering networks are not able to represent 
complex, edge-like patterns, e.g. checker-boards patterns.
Later, it was shown in~\cite{oyallon2017scaling,oyallon2018scattering} that hybrids of scattering networks and deep CNNs  can compete with end-to-end trained fully convolutional deep CNNs and outperform them in the small data regime. The authors replace early trainable convolutional layers with pre-defined scattering layers.
Furthermore, Cotter and Kinsbury~\cite{cotter2019learnable} design another hybrid version by using a scattering transform followed by a trainable 1d convolution as a building block of their scattering representation.
The latest work on scattering networks enables parametrization of the Morlet wavelets~\cite{gauthier2022parametric}, i.e. parameters are trainable and hence can be optimized for a suitable task. This results in a more problem adapted wavelet filter bank which is not necessarily a wavelet frame but is experimentally proven to be as stable to deformations (shear, scaling, and rotations) as classic scattering networks.

The importance of scattering networks lies in the fact that they represent the simplest non-trainable and nonlinear feature extractor which arranges convolutional filters and non-linearities in cascades. This constitutes the closest well-studied and mathematically sound proxy model~\cite{mallat2016understanding, wiatowski15} for trainable neural networks (e.g. CNNs) which are nowadays commonly used in a wide range of applications.

\subsection{Riesz transform and monogenic signal} 
In 1D, the analytic signal can be decomposed into amplitude and phase by using the Hilbert transform.
This decomposition is useful because it represents an orthogonal decomposition of signal information: amplitude refers to energetic information, while phase contains structural information.
The monogenic signal~\cite{felsberg01, felsbergPhD02}
is a high dimensional extension of the analytic signal, where the Hilbert transform is replaced by the Riesz transform, and inherits all of its relevant properties. 
The Riesz transform together with the Poisson kernel can be used to construct the so-called monogenic scale-space~\cite{felsberg04}.
Interestingly, the scale equivariance property of the Riesz transform has not been practically used 
until the works of Unser et al.~\cite{unser09, unser10} and Dobrovolskij et al. \cite{dobrovolskij19}. 

Since wavelets have been proven to be very useful for multiresolution signal analysis, it was natural to extend the concept of monogenic signal to wavelets.
This has been done through the generalization of the analytic wavelet transform to higher dimensions~\cite{olhede2009} by preserving the locality of the wavelet and introducing monogenicity through the Riesz transform.
Furthermore, in~\cite{held10, unser09}, the monogenic signal is applied to an isotropic wavelet that generates a wavelet frame. This results in a new mother wavelet that is directed, steerable, and also generates a wavelet frame. All these results enabled a multiresolution monogenic signal analysis.
In~\cite{unser10}, Unser and van de Ville derive an Nth order extension of the results in~\cite{held10} using higher order Riesz transforms. An example of this construction is illustrated on so called Riesz-Laplace wavelets. 
Unser and colleagues~\cite{unser10, unser09} are the first ones using the scale equivariance property of the Riesz transform, and have inspired the design of \textit{quasi monogenic shearlets}~\cite{hauser14}.

The so-called monogenic wavelet scattering network combines a scattering network with monogenic wavelets~\cite{chak2022}. However, scale equivariance is not used 
and translation equivariance is not assumed. Hence, it is a highly redundant and scale-dependent representation which extracts $90,000$ features for a $200 \times 200$ image.
The latest work~\cite{reinhardt2022rock} uses higher order Riesz transforms to classify rocks based on textures. However, here nonlinearity is not applied and no feature hierarchy has been constructed. Hence, the model resembles classical manual feature extractors. Nevertheless, this work serves as a proof of concept that with a relatively low number of Riesz-based features (between 100 and 400) good performance can still be achieved.

Lindeberg's paper~\cite{lindeberg2020} is the only work in literature that considers scale equivariance (or covariance) in the same context as we do: building a hierarchical nonlinear scale equivariant feature extractor. An oriented quasi quadrature measure is defined from first and second order Gaussian derivatives and used together with Gaussian derivatives and their absolute values for subsequent classification of textures. This representation depends on the sampling of the scale dimension which results in feature vectors of dimension 4,000. 
This feature dimension is higher than in the standard scattering networks~\cite{bruna2012-2}. However, it outperforms them on a more complicated texture classification problem on the KTH-Tips2b dataset proving the usefulness of the richer feature representation. The aim of our paper is to combine ideas from  \cite{lindeberg2020} and scattering networks \cite{bruna2012-2}. From \cite{lindeberg2020} we apply the idea of the construction of scale equivariant feature extraction, while from scattering network the idea of creating a nonexpansive feature extractor by using the amplitude of a complex filter as nonlinearity. The main difference is that the Riesz transform replaces Gaussian derivatives from \cite{lindeberg2020} or wavelets from \cite{bruna2012-2} for feature extraction.

Lindeberg's paper \cite{lindeberg2020} is significant as it represents one of the few works in theliterature that tackles the topic of scale generalization. Here, these methods are trained on a fixed range of scales and are able to generalize to a completely unseen range of scales not present in the training set. Works on Gaussian derivative networks \cite{lindeberg2022scale} and scale channel networks \cite{jansson2021exploring, jansson22} represent the extension of these ideas from scale space theory \cite{lindeberg1994scale, lindeberg1998feature, lindeberg2015image} in the deep learning framework. Riesz networks \cite{barisin22riesz} represent a related deep learning approach to scale generalization based on the Riesz transform. 
Finally, it is worth mentioning that the design of scale equivariant deep neural networks has recently gained attention in the deep learning community, e.g. see \cite{barisin22riesz, jansson22, lindeberg2022scale, sangalli2022scale,yang2022scale} to name a few.

\section{The Riesz transform}
\label{section:rt:steer}
In this section we formally introduce the Riesz transform and discuss its properties that will be needed for the construction of the Riesz feature representation in Section~\ref{chapter:Riesz:feature:representation}.

Let $L_2(\R^d) = \{ f \in \mathscr{L}: \int_{\R^d}{|f(x)|^2dx < \infty}\}$ be the square integrable functions from $\mathscr{L}$.
Formally, for a $d$-dimensional signal $f\in L_2(\R^d)$ (i.e. an image or a feature map), the Riesz transform of first order $\mathcal{R}=(\mathcal{R}_1,\cdots,\mathcal{R}_d)$ is defined in the spatial domain as $\mathcal{R}_{j}: L_2(\R^d) \to L_2(\R^d)$ for $j\in\{1,\cdots,d\}$
 \begin{equation*}
     \mathcal{R}_{j}(f)(x) = C_d \lim_{\epsilon \to 0}{\int_{\R^d \setminus B_{\epsilon}}{\frac{y_jf(x-y)}{|y|^{d+1}}dy}}, 
 \end{equation*}
where $C_d = \Gamma((d+1)/2)/\pi^{(d+1)/2}$ is a normalizing constant and $B_{\epsilon} \subset \R^d$ is ball of radius $\epsilon$ centered at the origin. Alternatively, the Riesz transform can be defined in the frequency domain via the Fourier transform $\mathcal{F}$
\begin{equation}
     \mathcal{F}(\mathcal{R}_j(f))(u) = -i\frac{u_j}{|u|}\mathcal{F}(f)(u) =  \frac{1}{|u|}\mathcal{F}(\partial_j f)(u), \quad j \in \{1,\cdots,d\}.
\end{equation}
 Higher order Riesz transforms are defined by applying a sequence of first order Riesz transforms. 
For the rest of the paper, let $n=(n_1,\cdots,n_d)$ be a multivector s.t. $|n|=\sum_{i=1}^d n_i =N \in \N$. 
We set

\begin{equation} \mathcal{R}^{(n_1,n_2,...,n_d)} (f)(x) := \mathcal{R}^{n_1}_1( \mathcal{R}^{n_2}_2( \cdots (\mathcal{R}^{n_d}_d(f)))(x),
\end{equation}
where $\mathcal{R}_j^{n_j}$ refers to applying the Riesz transform $\mathcal{R}_j$   $n_j$ times in a sequence. 
We refer to this as an $N$-th order Riesz transform.
The Riesz transform kernels of first and second order resemble those of the corresponding Gaussian derivatives. This can be explained by the relations
\begin{eqnarray}
     &&\mathcal{R}(f) =(-1)(-\triangle)^{-1/2} \nabla f \label{eq:riesz-gaussian-1st}\\ 
     &&\mathcal{R}^{(n_1,n_2,...,n_d)} (f)(x)= (-1)^N(- \triangle)^{-N/2} \frac{\partial^N f(x)}{\partial^{n_1}x_1 \cdots \partial^{n_d}x_d}.
     \label{eq:riesz-gaussian-higher}
\end{eqnarray}
The fractional Laplace operator $\triangle^{N/2}$ acts as an isotropic low-pass filter.
\subsection{Properties of the Riesz transform}
\label{properties:subsec}
The main properties of the Riesz transform can be summarized in the following way~\cite{unser10}:
\begin{itemize}
 \item \textbf{translation equivariance:} For $x_0 \in \R^d$,  and the translation operator $\mathcal{T}_{x_0}(f)(x):L_2(\R^d) \to L_2(\R^d)$  defined as $\mathcal{T}_{x_0}(f)(x) = f(x-x_0)$, we have
       $$ \mathcal{R}_j(\mathcal{T}_{x_0}(f))(x) = \mathcal{T}_{x_0}(\mathcal{R}_j(f))(x).$$

     \item \textbf{scale (dilation) equivariance:}
    For $a>0$ and a dilation or rescaling operator $L_{a}:L_2(\R^d) \to L_2(\R^d)$ defined as 
       $L_{a}(f)(x) = f(xa^{-1})$,
       scale equivariance holds
       $$ \mathcal{R}_j(L_{a}(f))(x) =  L_{a}(\mathcal{R}_j(f))(x).$$
       That is, the Riesz transform does not only commute with translations but also with scaling.
       
 \item \textbf{steerability:}
         Here, steering refers to the process of calculating an arbitrarily oriented filter as a linear combination of basis filters~\cite{Freeman93808}.
        The Riesz transform is defined only for the (orthogonal) base axes of $\R^n$. However, in practice the relevant features can be oriented arbitrarily. Hence, it is beneficial to orient the Riesz transform. This is achieved through the directional Hilbert transform $\mathcal{H}_v:L_2(\R^d) \to L_2(\R^d)$ in direction $v\in\R^d$, $||v||=1$ which is defined as $\mathcal{F}(\mathcal{H}_v(f))(u) = i  \sign(\langle u,v\rangle)$. 
       $\mathcal{H}_v$ is steerable in terms of the Riesz transform, that is 
       $$\mathcal{H}_v(f)(x) = \sum_{j=1}^d v_j \mathcal{R}_j(f)(x)  = \langle \mathcal{R} (f)(x), v\rangle.$$
       In 2d, the directional Hilbert transform for a unit vector $v=(\cos \phi, \sin\phi)$, $\phi\in\left[0,2\pi\right]$ becomes 
        $\mathcal{H}_v(f)(x) = \cos\phi\mathcal{R}_1(f)(x) + \sin\phi\mathcal{R}_2(f)(x)$.
        This is equivalent to the link between gradient and directional derivatives~\cite{unser10} and a very useful property for learning oriented features.
    
        \textbf{Higher order steerability:} similarly, higher order directional Hilbert transforms $\mathcal{H}_v^{(N)}:L_2(\R^d) \to L_2(\R^d)$ can be steered by higher order Riesz transforms:
        $$
        \mathcal{H}_v^{(N)}(f) (x) = \sum_{|n| = N} \frac{N!}{n!} v^n  \mathcal{R}^n(f)(x).
        $$
        where
        $v =(v_1, \cdots, v_d) \in \R^d$ is a unit vector  and $v^n:=\prod _{i=1}^N (v_i)^{n_i}$. For example, for $d=N=2$ and $v=(\cos{\phi},\sin{\phi})$ we have:
        $$ \mathcal{H}_v^{(2)}(f)(x) = \cos^2(\phi)\mathcal{R}^{(2,0)}(f)(x) + \sin^2(\phi)\mathcal{R}^{(0,2)}(f)(x) + 2 \cos(\phi)\sin(\phi)\mathcal{R}^{(1,1)}(f)(x).$$
 \item \textbf{all-pass filter~\cite{felsberg01}:}
    Let $H = (H_1, \cdots, H_d) =  (i\frac{u_1}{|u|}, \ldots, i\frac{u_d}{|u|})$ be the Fourier transform of the Riesz kernel. 
    The energy of the Riesz transform for frequency $u \in \R^d$ is defined as the norm of the $d$-dimensional vector $H(u)$. It is constant to one, i.e., 
       $|H(u)| = 1$  for all $u \neq 0.$
       The all-pass filter property reflects the fact that the Riesz transform is a non-local operator and that every frequency is treated fairly and equally. Combined with scale equivariance, this property guarantees that the Riesz transform is indeed able to extract useful features independent of the scale through the use of a single operator.
\end{itemize}

  \section{Riesz transform and signal decomposition}
This section discusses signal decomposition using the Riesz transform. 
This can be done through the monogenic signal using the first order Riesz transform (Section~\ref{sec:monogenic:signal}) or alternatively through higher order Riesz transforms (Section~\ref{subsect:decomp:high}). This section is needed to analyze the nonexpansiveness properties of the Riesz feature representation in Section~\ref{sec:path:ordered:scatter}. 

\subsection{Decomposition using the monogenic signal}
\label{sec:monogenic:signal}
Generally, the goal of the analytic or the monogenic signal is to extract useful information from 
a signal by means of decomposition. For example, a 2d signal $f$ can be decomposed into local amplitude, orientation, and phase. Here, we summarize results for the 2d case from~\cite{felsberg01}.
First, the monogenic representation $ f_{\mathcal{M}}(x)$ of the signal $f\in L_2(\R^2)$ at point $x \in \R^2$  is calculated:
\begin{equation}
\label{eq:monogenic:signal}
    f_{\mathcal{M}}(x) = f(x) + i\mathcal{R}_1 f(x) + j \mathcal{R}_2 f(x) 
\end{equation}
where $i$ and $j$ can be seen as imaginary units and are related to the notation and calculus from Clifford analysis~\cite{felsberg01}. 

The local amplitude $|f_{\mathcal{M}}(x)|$ is given as $L_2$-norm of $f_{\mathcal{M}}(x)$, i.e. \\
$ |f_{\mathcal{M}}(x)| = \sqrt{f(x)^2 + (\mathcal{R}_1 f(x))^2 + (\mathcal{R}_2 f(x))^2}.$
Local orientation and local phase refer to structural information at the point $x$. Local orientation is derived from $(\mathcal{R}_1 f(x),\mathcal{R}_2 f(x))$ following the differential interpretation of the Riesz transform:
$v(x) = \tan^{-1}\Big(\frac{\mathcal{R}_2 f(x)}{\mathcal{R}_1 f(x)}\Big). $
Local phase is defined in the direction $f_D(x) =(-\mathcal{R}_2 f(x),\mathcal{R}_1 f(x),0)$ that is orthogonal to the local orientation by
$ \theta(x) = \frac{f_D(x)}{|f_D(x)|}\tan^{-1}\Big(\frac{\sqrt{(\mathcal{R}_1 f(x))^2 + (\mathcal{R}_2 f(x))^2}}{f(x)}\Big).$ 
Given local phase and local amplitude of the monogenic signal, the full signal can be reconstructed:
$$ f_{\mathcal{M}}(x) = |f_{\mathcal{M}}(x)|e^{(-j,i,0) \theta(x)}.$$

\subsection{Decomposition using higher order Riesz transforms}
\label{subsect:decomp:high}
Higher order Riesz transforms also enable a useful signal decomposition. Here, we follow the results from~\cite{unser09}. The first theorem states the invertibility of higher order Riesz transforms. 
\begin{theorem}
The N-th order Riesz transform achieves the following decomposition:
\begin{equation}
    \sum_{|n| = N}\frac{N!}{n!}(\mathcal{R}^{n_1}_{1} \cdots \mathcal{R}^{n_d}_{d})^*(\mathcal{R}^{n_1}_{1} \cdots \mathcal{R}^{n_d}_{d}) = \text{Id},
\end{equation}
\label{thm:1}
\end{theorem}
A proof is given in~\cite{unser10}.
Hence, the N-th order Riesz transform decomposes a signal into $p(N,d) = {N+d-1 \choose d-1}$ components from which the original signal can be reconstructed.
The next theorem states that this decomposition perfectly preserves the $L_2$-norm of the signal. 
\begin{theorem}
\label{parseval:thm}
The N-th order Riesz transform satisfies the following Parseval-like identity:
\begin{equation}
    \sum_{|n| = N} \frac{N!}{n!}\langle \mathcal{R}^nf , \mathcal{R}^ng \rangle_{L_2} = \langle f,g \rangle_{L_2}
\end{equation}
for every $f,g \in L_2(\R^d).$
\end{theorem}
This implies the conservation of signal energy $||f||$
\begin{equation}
    \sum_{|n| = N} \frac{N!}{n!} || \mathcal{R}^nf||^2 = ||f||^2.
\end{equation}
and the contraction property
\begin{equation}
||\mathcal{R}^nf||^2 \leq ||f||^2.
\end{equation}
For the case $N = 1$, this turns out to be $||f||^2 = ||\mathcal{R}_1(f)||^2+||\mathcal{R}_2(f)||^2$, 
or equivalently in terms of equation (\ref{eq:monogenic:signal}): $||f_{\mathcal{M}}||^2 =2 ||f||^2.$

\subsubsection{Nonexpansiveness of the Riesz transform:} Due to the linearity of the Riesz transform, we have 
\begin{equation}
||\mathcal{R}^n(f)-\mathcal{R}^n(g)||^2 = ||\mathcal{R}^n(f-g)||^2\leq ||f-g||^2.
\end{equation}
This property will be useful for understanding the decay of energy (amplitude) with an increase in the depth of the Riesz hierarchical representation.

\section{Riesz hierarchical representation}
\label{chapter:Riesz:feature:representation}
The construction of the Riesz hierarchical network is similar to that of scattering networks~\cite{mallat2012} (Appendix~\ref{appendix:scattering}). The construction can be summarized as a three step process: apply a complex filter bank to the image, apply a modulus operator as non-linearity, and repeat the previous two steps to construct hierarchies. 
The main novelty is in fact that we use a smaller and less redundant filter bank compared to~\cite{bruna2012-2} 
based on the Riesz transform which ensures stability to variations in scales. The properties of the Riesz transform as stated above ensure that the most important properties of scattering networks such as nonexpansiveness and translation equivariance are preserved. 

\subsection{Base function and its properties}
\label{base-function5:1}
Instead of the Morlet wavelet, we use a scale-free Riesz transform kernel to construct a complex base function and eliminate the need for scale selection.
For $x=(x_1,x_2) \in \R^2$, the complex base function $ \psi:\R^2 \to \C$ is defined as 
\begin{equation}
    \label{eq:base:function}
    \psi (x) =  i r_1(x) +  r^{(2,0)}(x),
\end{equation} 
where $r_1$ and $r^{(2,0)}$ are kernels of the Riesz transform of first order $\mathcal{R}_1$ and second order $\mathcal{R}^{(2,0)}$, respectively.
In other words, for $f \in L_2(\R^2)$, we have
$$ (f\ast \psi)(x) = i \mathcal{R}_1(f)(x) +   \mathcal{R}^{(2,0)}(f)(x)\ .$$ 
Note that convolution with $\psi$ is \textbf{scale equivariant} and a \textbf{quadrature filter}. 
The latter is easily deduced from its \textbf{differential interpretation} since the first (second) order Riesz transform resembles the first (second) order derivative operator 
(see Equations~(\ref{eq:riesz-gaussian-1st}) and (\ref{eq:riesz-gaussian-higher})) and can be seen as an edge (line) detector.

\textbf{Steerability:}
 An image can be thought of as a composition of differently oriented components, signals, or patterns. 
 Note that the base function $\psi$ is not an isotropic function as it is oriented vertically. 
Hence, the convolution with the base function preserves (amplifies) all patterns rotated in (or close to) the vertical axis, while it diminishes signals oriented horizontally. For this reason, it is useful to define a rotated version of the base function $\psi$. This is particularly important for images that are not uni-directional, i.e. that have multiple relevant orientations.
 Our base function and the rotated filters derived from it are steerable, i.e. there exist basis filters from which any rotated filter can be steered. 

Formally, let $G$ be a finite rotation group.
For group element $r= (\cos \phi, \sin \phi) \in G$, the base function $\psi$ can be rotated using the steerability of the Riesz transform via
$$ \psi_r (x) : = \psi(r^{-1}x) = i h_r(x) +  h_r^{(2)}(x)$$
where $h_r$ and $h^{(2)}_r$ are kernels of the directional Hilbert transform or the steered Riesz transforms of first order ($\mathcal{H}_r$) and second order ($\mathcal{H}^{(2)}_r$). 
This yields $$ (f\ast \psi_r)(x) = i \mathcal{H}_r(f)(x) + \mathcal{H}^{(2)}_r(f)(x),$$ where
$$\mathcal{H}_r (f) (x) = \cos\phi \mathcal{R}_1(f)(x) + \sin \phi \mathcal{R}_2(f)(x),$$
$$ \mathcal{H}^{(2)}_r(f)(x) = \cos^2(\phi)\mathcal{R}^{(2,0)}(f)(x) + \sin^2(\phi)\mathcal{R}^{(0,2)}(f)(x) + 2 \cos(\phi)\sin(\phi)\mathcal{R}^{(1,1)}(f)(x).$$
Hence, steering this type of function into an arbitrary number of orientations only requires computation of five Riesz transforms: two of the first order ($\mathcal{R}_1,  \mathcal{R}_2$) and three of the second order ($\mathcal{R}^{(2,0)}, \mathcal{R}^{(0,2)}, \mathcal{R}^{(1,1)}$). 
A visualization of the resulting Riesz transform features can be found in Figure~\ref{fig:riesz:transform:mnist} of Appendix~\ref{appendix:visualization}.

\textbf{Nonexpansiveness of the base function $\psi$:} 
Nonexpansiveness is preserved if the complex base function is multiplied by a real scalar $C \leq \frac{1}{2}$. Generally, it holds that
$$ ||  C\cdot(f\ast \psi) -  C\cdot(g\ast \psi) ||^2 \leq C || (f-g)\ast \psi ||^2 = C||\mathcal{R}_1(f-g)||^2+ C||\mathcal{R}^{(2,0)}(f-g)||^2 \leq 2C ||f-g||^2.$$
Similarly, this holds for the rotated version of $\psi_r$ which follows from Lemma~\ref{two:ineq:lemma} in Appendix~\ref{appendix:technical}. 
The reason for requiring nonexpansiveness in~\cite{mallat2012} is that the representation should be stable (Lipschitz continuous) to small deformations, i.e. it should not amplify their effect~\cite{mallat2016understanding}.
Note that the selection of the coefficient $C$ determines the decay of energy as we increase the depth of the Riesz representation. Its role in the Riesz representation will be analyzed later.

 \textbf{Zero integral of base function $\psi$:} 
Lemma~\ref{lemma:zero:int} in Appendix~\ref{appendix:technical} shows that $$\text{p.v.} \int_{\R^2} \psi(x) dx = 0\ ,$$
where p.v. denotes the Cauchy principal integral (note that the convolution with the Riesz kernel is not defined at 0).
This result serves as a motivation to introduce non-linearity in the scenario where we assume (global) translation invariance. If one applies global average pooling without non-linearity, the result would inevitably be $0$ for every feature. This in turn would remove all the discriminative characteristics of the transformed signal. Additionally, the zero integral property guarantees that the representation is invariant to constant shifts in gray values.

\subsection{Path ordered scattering}
\label{sec:path:ordered:scatter}
\label{subsection:path:order:scatter}


  The goal of this section is to design a universal hierarchical representation of the image structure which is useful in the sense that a simple classifier (e.g. PCA or SVM) can be constructed on top of the representation without the need to train a deep representation or adjust parameters.
  
  \subsubsection{Alternative to Scattering networks}
 Let $G_M=\{k\frac{\pi}{M}, k\in\{0, 1,\cdots,M-1\}\}$ be a finite rotation group  where $M\in \N$ controls the discretization of the group. 
  Then we define the base transformation layer by
  $$W(f) := \{ C\cdot(f\ast\psi_r), r \in G_M\}$$
  where $C\le \frac 1 2$ is a scaling constant which is needed to preserve nonexpansiveness or control energy decay. To ensure that both the coordinate axes and the diagonal directions are included in $G_M$, we choose $M=4m$ for some $m \in \N$.

  Similar to scattering networks, we then apply a non-linearity operator. Here, we apply the amplitude operator to every feature map from $W(f)$
 and discard the phase. This yields
  $$ S(f) = A(W(f)):= \{ A(g), g \in W(f) \},$$
  where $A:L^2(\R^d, \C) \to L^2(\R^d)$ is the pointwise amplitude operator on complex functions, i.e. $A(g_1(x)+i g_2(x)) = \sqrt{g_1(x)^2+g_2(x)^2}.$
  To create a multilayer deep representation, we apply the operator $S$ sequentially. That is, for the $k$-th layer, we have
  \begin{equation}
    \label{eq:Sk}
    S_k(f) := S^k(f) = S(\cdots S(f)).
  \end{equation}
  and let $S_0 := f.$
  Selected feature maps from $S_k$ up to depth $3$ are visualized in Figure~\ref{riesz:representation:kth} in Appendix~\ref{appendix:visualization}.
  The final representation $\Phi$ with $K\in \N$ layers is the ordered list of feature maps from all depths 
  \begin{equation}
      \label{eq:Phi}
      \Phi(f) = \Big( S_k(f) \quad | \quad k=0, \cdots, K  \Big),
  \end{equation}
  This representation yields in total $\sum_{k=0}^K M^k$ features per pixel. 
  
  \textbf{Nonexpansiveness of the Riesz representation $\Phi(f)$:}
  A requirement from~\cite{bruna2012-2,mallat2012} is that the total energy of the layer equals the energy of the input $f$. Although this is needed to achieve Lipschitz continuity to small deformations, it results in fast energy decay and reduces the importance of features in deeper layers. For this reason the depth of scattering networks is usually limited to 2~\cite{mallat2012,bruna2012-2}.
  However, this property is only relevant when using classifiers such as PCA that do not use any feature normalization or preprocessing.
  On the other hand, prior to training a SVM on the features from the Riesz representation, the feature vectors are normalized coordinatewise with maximal absolute value.
  This normalization removes the effect of energy decay. 

  Strictly theoretically, to achieve nonexpansiveness of $\frac{1}{K+1}\Phi$ for $M=4m$ for $m\in \N$, 
  one can choose $C = \frac{1}{M}$ by the following arguments.
  
  First, we analyze nonexpansiveness of the operator $S(f) = A(W(f))$. Here, we have   \begin{align*}
      ||S(f) -S(g)||^2 &= ||A(W(f)) - A(W(g))||^2 \leq || A\big(W(f)-W(g)\big) ||^2 =
      \\
     &= ||W(f-g)||^2 
      = C \Big( \sum_{r \in G_M} ||i \cdot\mathcal{H}_r(f-g)+\mathcal{H}_r^{(2)}(f-g)||^2 \Big) =
      \\
      &= C \Big( \sum_{r \in G_M} ||\mathcal{H}_r(f-g)||^2 + \sum_{r \in G_M}||\mathcal{H}_r^{(2)}(f-g)||^2 \Big).
  \end{align*}
  
  From Lemma \ref{two:ineq:lemma} we have $\sum_{r \in G_M} ||\mathcal{H}_r(f-g)||^2 \leq \frac{M}{2}||f-g||$. A similar statement can be shown for the second summand.

For $C = \frac{1}{M}$, this yields 
$$
||S(f) -S(g)||^2 \le C \left(\frac M 2 ||f-g||^2+  \frac M 2 ||f-g||^2 \right) =||f-g||^2.
 $$

 Since $\Phi$ defined for some $K\in \N$ has depth $K$, it needs to be scaled by $\frac{1}{K+1}$ to achieve nonexpansiveness.
  We experimentally investigate the effect of the constant $C$ on the overall representation performance in Section~\ref{sec:nonexpand:kth}.
  
  \subsubsection{Equivariance properties}
   As stated in Section~\ref{section:rt:steer} above, the Riesz transform is equivariant to two types of transformations: translation and scaling. In this section, we show that the Riesz representation inherits these equivariances. To prove this, it is enough to prove that the operator $S$ is equivariant to these transformations since the Riesz representation consists of applying this operator in sequence. The operator $S$ consists of two types of transformations: a convolution with complex base functions $\psi$ and applying the pointwise amplitude operator $A$. By construction the base functions preserve both types of equivariances. The amplitude operator $A$ is a pointwise function and hence preserves equivariance to scaling and translation. 
   Next, we sketch the proof of the last claim.
   \begin{theorem}
           Let  $A: L_2(\R^d, \C)\to L_2(\R^d)$ be a (continuous) pointwise operator that only depends on the pixel values. Then $A$ is equivariant to translation and scaling.
   \end{theorem}
   \begin{proof}
    Being a pointwise operator formally means there exists a function $F:\C\to\R$ 
    such that for every $x\in\R^d$ it holds that $A(f)(x) = F(f(x))$.
   For a translation operator $T_{x_0}: L_2(\R^d,\C) \to L_2(\R^d, \C)$ for $x_0 \in \R^d$ defined as $T_{x_0}(f)(x) = f(x-x_0)$, it follows
    $$T_{x_0}(A(f))(x) = A(f)(x-x_0)
    = F(f(x-x_0)) = F(T_{x_0}(f)(x)) = A(T_{x_0}(f))(x).$$
   For a scaling operator $L_a: L_2(\R^d,\C) \to L_2(\R^d,\C)$ where $a>0$ defined as $L_a(f)(x) = f(xa^{-1})$, the proof is analogous:
   $$ L_a(A(f))(x) = A(f)(xa^{-1}) = F(f(xa^{-1})) = F(L_a(f)(x)) = A(L_a(f))(x).$$
   \end{proof}

  \subsection{Pooling operations for scale and translation invariance}
  \label{sec:pooling}
  The output of the operator $\Phi$ from Equation (\ref{eq:Phi}) is a sequence of feature maps which have the same size as the input image. 
  Furthermore, the operator $\Phi$ is both scale and translation equivariant. 
  The goal is to use a pooling operation $\rho : L_2(\R^d) \to \R$ that enables the transition from equivariance to invariance. 
  
  For that purpose, a global pooling operator $\rho$ is applied to every feature map from $\Phi$. A global pooling operator is an operator which takes an image or feature map $f\in L_2(\R^d)$ as an input and computes a single summary statistic on the whole image, i.e. $\rho: L_2(\R^d) \to \R$. 
  When applied to an image, this pooling function $\rho$ becomes a global operator over a bounded, discrete domain.
  
  In contrast, one can compute summary statistics on non-overlapping windows on the image. These pooling statistics depend on the position in the image or local image structure. Hence, they are called localized or local pooling operators. These operators only guarantee local invariance depending on the size of the window on which they are calculated. The drawback with local pooling operators is that they are not comparable for images of different sizes since the feature vectors after pooling have different sizes. Hence, we focus on global pooling on bounded discrete domains $D\subset \Z^d$, only.

  Standard choices for global pooling operators are average and maximum pooling. Average pooling takes the mean value of the feature map. To obtain comparable values for different scales, it is important that the ratio between the object and the background is fixed when altering the scale. 
   An alternative is global max pooling which reports the maximal value of the feature map. However, in noisy conditions the extreme values are more sensitive to outliers compared to the mean values. Furthermore, this operator does not guarantee nonexpansiveness of the Riesz feature representation.
   Hence, we use global average pooling as pooling operator $\rho$ throughout the rest of the paper.


\section{Experiments}

Throughout this section, we use a Riesz feature representation with depth 3 based on a discrete rotation group with $M=4$ angles, if not specified otherwise. In this case, the output of the Riesz feature representation for an image of arbitrary size consists of 85 features. 
The nonexpansiveness constant $C$ from Section~\ref{subsection:path:order:scatter} is set to $1$, as justified later experimentally in Section~\ref{sec:nonexpand:kth}.

\subsection{MNIST Large Scale}
\label{mnist:exp}
\begin{figure}[h]
    \centering
    \includegraphics[width = 0.18 \textwidth]{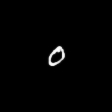}
    \includegraphics[width = 0.18 \textwidth]{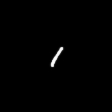}
    \includegraphics[width = 0.18 \textwidth]{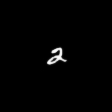}
    \includegraphics[width = 0.18 \textwidth]{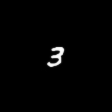}
    \includegraphics[width = 0.18 \textwidth]{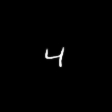} 
    \hfill
    \\
    \includegraphics[width = 0.18 \textwidth]{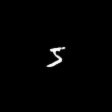}
    \includegraphics[width = 0.18 \textwidth]{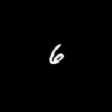}
    \includegraphics[width = 0.18 \textwidth]{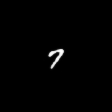}
    \includegraphics[width = 0.18 \textwidth]{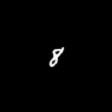}
    \includegraphics[width = 0.18 \textwidth]{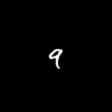}
    \hfill
    \\
    \caption{The 10 classes of the MNIST Large Scale data set. All images have $112 \times 112$ pixels.}
    \label{fig:mnist-classes}
\end{figure}

\begin{figure}[h]
    \centering
    \includegraphics[width = 0.18 \textwidth]{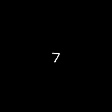}
    \includegraphics[width = 0.18 \textwidth]{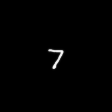}
    \includegraphics[width = 0.18 \textwidth]{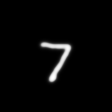}
    \includegraphics[width = 0.18 \textwidth]{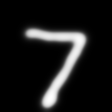}
    \includegraphics[width = 0.18 \textwidth]{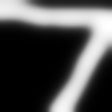}
    \caption{Variation of scales in the MNIST Large Scale data set (from left to right): scales 0.5, 1, 2, 4, and 8. All images have $112 \times 112$ pixels.}
    \label{fig:mnist-scales}
\end{figure}

The MNIST Large Scale 
 data set~\cite{jansson22} derived from the MNIST data set~\cite{lecun98} consists of images of digits between 0 and 9 each digit forming one of ten classes (Figure~\ref{fig:mnist-classes}). The images are rescaled to a wide range of scales to test scale generalization abilities of classifiers (Figure~\ref{fig:mnist-scales}).
 The training set has 50,000 images of the single scale $1$. The test set consists of scales ranging in $\left[0.5,8\right]$ with 10,000 images per scale. All images consist of $112\times 112$ pixels.
 
\begin{figure}[]
    \centering
    \includegraphics[scale = 1]{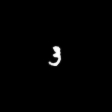}
    \includegraphics[scale = 4]{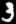}
    \caption{Extracting the bounding box (right) from the input image (left). The input image consists of $112\times112$ pixels, the bounding box of $16\times24$ pixels.}
    \label{fig:mnist-boundbox}
\end{figure}

\begin{table*}[]
    \centering
     \scalebox{0.9}{
    \begin{tabular}{|c|c|c|c|c|c|c|c|c|c|}
    \hline
     \multicolumn{2}{|c|}{scale} & 0.5 & 0.595 & 0.707 & 0.841 & 1 & 1.189 & 1.414 & 1.682 \\
    \hline
     \multicolumn{2}{|c|}{CNN tr1~\cite{jansson22}} & 61.84 & 85.3 & 96.10 &  98.73 &  99.32 & 98.50  & 85.36 &  52.61
     \\
    \hline
    \multicolumn{2}{|c|}{FovAvg 17ch tr1~\cite{jansson22}} & 98.58 & 99.05 & \textbf{99.33} & \textbf{99.39} & \textbf{99.40} & \textbf{99.39} & \textbf{99.38} & \textbf{99.36}
    \\
    \multicolumn{2}{|c|}{FovMax 17ch tr1~\cite{jansson22}} & \textbf{98.71} & \textbf{99.07} & 99.27 & 99.34 & 99.37 & 99.35 & 99.36 & 99.34 
    \\
    \hline
    \multicolumn{2}{|c|}{RieszNet~\cite{barisin22riesz}} & 96.34 & 97.59 & 98.06 & 98.54 &  98.58 & 98.50 & 98.45  & 98.40 
    \\
    \multicolumn{2}{|c|}{RieszNet-pad40~\cite{barisin22riesz}} & 96.34 & 97.55 & 98.07 & 98.47 & 98.63  & 98.58 & 98.53  & 98.44 
    \\
    \hline
    \multicolumn{2}{|c|}{ours-MLP} & 82.78 & 91.47 & 94.19 & 95.23 & 96.23 & 95.73 & 95.79 & 95.63 
    \\
    \hline
     \multicolumn{2}{|c|}{ours-SVM} & 82.36 & 91.29 & 93.86 & 94.83 & 95.74 & 95.19 & 95.13 & 95.04 
     \\
    \hline 
    \multicolumn{10}{c}{} \\
    \hline
    scale & 2 & 2.378 & 2.828 & 3.364 & 4 & 4.757 & 5.657 & 6.727 & 8 \\
      \hline
     CNN tr1~\cite{jansson22} & 36.82 & 28.55 & 22.38 & 19.04 & 14.47 & 11.71 & 11.50 & 10.88 & 10.68
     \\
    \hline
    FovAvg 17ch tr1~\cite{jansson22} & \textbf{99.35} & 99.31 & 99.22 & 99.12 & 98.94 & 98.47 & 96.20 & 89.17 & 71.31
    \\
    FovMax 17ch tr1~\cite{jansson22}  & 99.33 & \textbf{99.35} & \textbf{99.34} & \textbf{99.35} & \textbf{99.34} & \textbf{99.27} & \textbf{97.88} & 92.76 & 79.23 
    \\
    \hline
    RieszNet~\cite{barisin22riesz} & 98.39 & 98.24 & 98.01 & 97.51 & 96.42 & 93.50 & 81.58 & 67.66 & 51.82
    \\
    RieszNet-pad40~\cite{barisin22riesz}  &  98.46 & 98.39 & 98.34 & 98.29 & 98.16 & 97.80  & 96.82 &  \textbf{93.75} & \textbf{83.6}  \\
    \hline
    ours-MLP & 95.60 & 95.73 & 95.79 & 95.69 & 95.61 & 95.64  & 94.76 & 88.26 & 67.92 
    \\
    \hline
    ours-SVM  & 95.08 & 95.02 & 95.01 & 94.97 & 95.05 & 95.03 & 94.00 & 86.19 & 63.96\\
    \hline
    \end{tabular}
    }
    \caption{Classification accuracies (in \%) for the MNIST Large Scale data set. Several methods trained on the full training set (50,000 images) at scale 1. Best performing method bold. Accuracies for CNN tr1, FovAvg 17ch tr1 and FovMax 17ch tr1 are taken from~\cite{ jansson2021exploring,jansson22}.} 
    \label{fig:mnist-scale-generalization}
\end{table*}

 \textbf{Why is scale equivariance not perfectly suited for this or similar problems?} 
 Remember that scale equivariance implies that the Riesz transform commutes with the scaling operator. In other words, it does not matter in which order we apply Riesz transform and scaling operator to the image. 
 In practice, however, the images under consideration have a bounded domain. Hence, mean pooling as discussed above may be sensitive to changes of the size ratio between object and background. Also, problems may arise when an object partially leaves the image when upscaling.
 This problem is particularly relevant for non-local feature extractors such as the Riesz transform. It is interesting to notice that neural networks might overcome this issue through the training procedure as demonstrated for a trainable Riesz network in~\cite{barisin22riesz}.

\begin{table*}[h]
    \centering
     \scalebox{0.87}{
    \begin{tabular}{|c|c|c|c|c|c|c|c|c|c|}
    \hline
     \multicolumn{2}{|c|}{scale} & 0.5 & 0.595 & 0.707 & 0.841 & 1 & 1.189 & 1.414 & 1.682
     \\
    \hline
    \multicolumn{2}{|c|}{RieszNet train 1,000~\cite{barisin22riesz}} & \textbf{86.94} & \textbf{89.07} & \textbf{90.98} & \textbf{91.54} & \textbf{91.64} & \textbf{91.93} & \textbf{91.61} &  \textbf{91.42} 
    \\
     \hline
    \multicolumn{2}{|c|}{ours-MLP train 1,000} & 70.72 &  80.65 & 84.38 & 87.13 & 89.00 & 87.21 & 87.53 & 87.26 
    \\
     \multicolumn{2}{|c|}{ours-SVM train 1,000} & 71.16 & 80.28 & 82.98 & 84.79 & 87.49 & 85.09 & 84.99 & 84.81 
     \\
     \hline
     \multicolumn{2}{|c|}{ours-SVM train 5,000} & 78.02 & 86.97 & 89.81 & 91.20 & 92.26 & 91.48 & 91.31 & 91.11 
     \\
     \multicolumn{2}{|c|}{ours-SVM train 20,000} & 80.38 & 89.67 & 92.20 & 93.31 & 94.66 & 93.91 & 93.97 & 93.70
     \\
     \multicolumn{2}{|c|}{ours-SVM train 50,000} & 82.36 & 91.29 & 93.86 & 94.83 & 95.74 & 95.19 & 95.13 & 95.04 
     \\
     \hline
     \multicolumn{2}{|c|}{ours-MLP train 50,000} & 82.78 & 91.47 & 94.19 & 95.23 & 96.23 & 95.73 & 95.79 & 95.63 \\
    \hline
    \multicolumn{10}{c}{} \\
    \hline
    scale & 2 & 2.378 & 2.828 & 3.364 & 4 & 4.757 & 5.657 & 6.727 & 8 \\
    \hline
    RieszNet train 1,000~\cite{barisin22riesz} & \textbf{90.93} & \textbf{90.24} & \textbf{89.32} & \textbf{87.97} & \textbf{85.78} & 82.01 & 74.84 & 67.31 & 56.88 
    \\
    \hline
    ours-MLP train 1,000 & 87.48 & 87.28 & 87.27 & 87.33 & 87.27 & \textbf{86.98} & \textbf{86.14} & \textbf{78.00} & \textbf{59.79}
    \\
    ours-SVM train 1,000 & 84.74 & 84.75 & 84.55 & 84.50 & 84.53 & 84.26 & 83.71 & 70.97 & 48.33
    \\
    \hline
    ours-SVM train 5,000 & 91.38 & 91.18 & 91.12 & 91.18 & 91.14 & 91.01 & 90.15 & 78.23 & 53.04 
    \\
    ours-SVM train 20,000 & 93.83 & 93.75 & 93.71 & 93.64 & 93.77 & 93.59 & 92.82 & 84.19 & 60.69
    \\
    ours-SVM train 50,000 & 95.08 & 95.02 & 95.01 & 94.97 & 95.05 & 95.03 & 94.00 & 86.19 & 63.96
    \\
    \hline
    ours-MLP train 50,000 & 95.60 & 95.73 & 95.79 & 95.69 & 95.61 & 95.64  & 94.76 & 88.26 & 67.92 
    \\
    \hline 
    \end{tabular}
    }
    \caption{Classification accuracies (in \%) for the MNIST Large Scale data set. Methods trained on a reduced training set comprising $1,000$ images. Best performing method in bold. Training set scale is $1$. Note that the authors in \cite{jansson22} perform experiments with a small number of samples as well but with training scales in the interval $[1,4]$ and testing on the whole range. This is a slightly different experiment design and hence was not reported in this table.
    }
    \label{fig:mnist-scale-generalization:small:datasets}
\end{table*}

To deal with the changing size ratio between the object and the background, one can determine the bounding box around the object of interest.
Restriction to the bounding box will preserve the size ratio between object and background. Note that in general, the bounding box computation should be scale equivariant to preserve the same property of the Riesz feature representation. Here, we design a simple scale equivariant four step bounding box algorithm: 
\begin{enumerate}
    \item normalize the image gray values with min-max normalization to be in the interval~$[0,1]$,
    \item pad the image with 50 zero pixels on each side,
    \item threshold the image with $t=0.5$, and \item draw the bounding box on the binary image.
\end{enumerate}

Finally, the bounding box is enlarged to capture black background by elongating the diagonals by $40\%$. The bounding box is cropped from the image (Figure~\ref{fig:mnist-boundbox}) and input to the Riesz feature representation. Afterwards, the SVM classifier is trained on the output from the previous step. Additionally, we train a multilayer perceptron (MLP) on the same output to compare the performance of these two classifiers. Details on the  MLP and a small ablation study on the depth and the rotation group of the Riesz feature representation are given in Appendix~\ref{appendix:mlp}. In this section, MLP refers to an MLP applied to a Riesz representation of depth 3 with 4 angles. It has roughly $30,000$ trainable parameters, as described in Appendix~\ref{appendix:mlp}.

The goal of this experiment is to validate scale equivariance properties of the Riesz representation as well as to deduce how many training images are needed to achieve a decent performance.  
We compare our method with several state-of-the-art methods: the RieszNet~\cite{barisin22riesz} and a CNN on rescaled versions of the images~\cite{jansson2021exploring, jansson22}. 
The RieszNet~\cite{barisin22riesz} is a deep neural network where spatial convolutions are replaced by the first and second order Riesz transforms while max pooling is completely avoided. This results in a fully trainable scale invariant neural network.
The RieszNet version used here has around $18,000$ parameters.
The second method called Foveated scale-channel networks
\cite{jansson2021exploring, jansson22} is a classic CNN with around $70,000$ parameters which is applied to rescaled versions of the images and finally (max or average) pooled across the scales. Here, three sampling factors $\{2,\sqrt{2}, \sqrt[4]{2}\}$ of the interval $[\frac{1}{2},8]$ are compared. We select the best performing one, i.e. the one for which 
17 scales in the interval $[\frac{1}{2},8]$ are sampled with step width $\sqrt[4]{2}$ and used for rescaling the images. 
This is in contrast to the Riesz-based methods where this step is completely avoided. Table~\ref{fig:mnist-scale-generalization} summarizes the results.

Generally, accuracies of our method are stable for scales in the range $[0.707, 5.657]$ for both, SVM and MLP. 
The MLP performs slightly better (around 1\%) than the SVM. However, the costs of training and parameter tuning for the MLP are higher.
Both competing methods achieve $4-5\%$ higher accuracy than our Riesz representation baseline with SVM.
However, this agrees with similar experiments reported in the literature~\cite{oyallon2015} when comparing scattering and deep networks on large datasets.

It is of interest to analyze the performance of the methods for smaller sizes of the training set \cite{jansson22}. The authors in \cite{jansson22} experiment with training sets in the scale $[1,4]$ with sizes as small as 100 images. Here, we decided to keep the same experiment design as above: train only on a single scale $1$. Accuracies depending on the number of training images are reported in Table~\ref{fig:mnist-scale-generalization:small:datasets}. 
Here, we compare with the RieszNet, only. 
With the SVM, the accuracy improves significantly with increased number of training images (by $10\%$ for 50,000 vs 1,000 training images) since more training images generally increase the diversity of handwriting styles for digits. However, the performance on the relatively small training set of 1,000 images (100 training images per class) is around $85\%$ for the SVM, and stable for the scales in the range $[0.707, 5.657]$. When trained on the same training set of size 1,000, the MLP is better than the SVM ($85\%$ vs $87\%$). Interestingly, the MLP outperforms the RieszNet for scales larger than $4.757$. This could be due to the window cropping procedure.

\begin{figure}
    \centering
    \includegraphics[width = 0.19\textwidth]{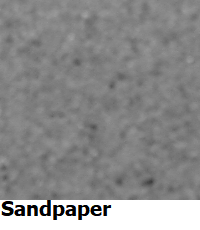}
    \includegraphics[width = 0.19\textwidth]{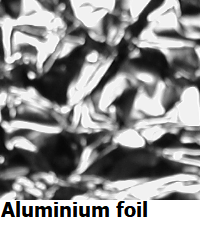}
    \includegraphics[width = 0.19\textwidth]{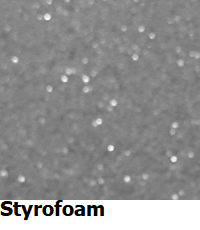}
    \includegraphics[width = 0.19\textwidth]{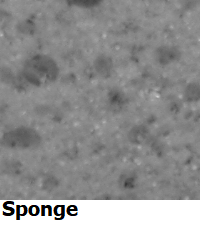}
    \includegraphics[width = 0.19\textwidth]{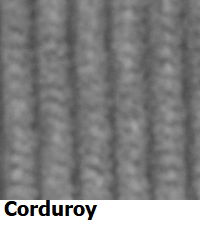}
    \includegraphics[width = 0.19\textwidth]{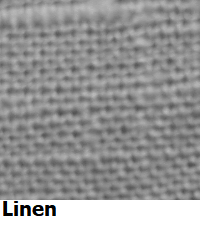}
    \includegraphics[width = 0.19\textwidth]{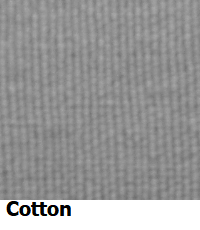}
    \includegraphics[width = 0.19\textwidth]{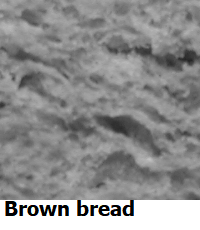}
    \includegraphics[width = 0.19\textwidth]{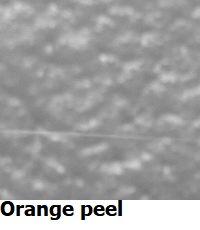}
    \includegraphics[width = 0.19\textwidth]{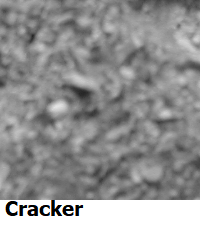}
    \caption{Sample classes in KTH-tips dataset.}
    \label{fig:kth-classes}
\end{figure}

\subsection{KTH-tips}
\begin{figure}
\vspace{0.1cm}
    \centering
    \includegraphics[width = 0.19\textwidth]{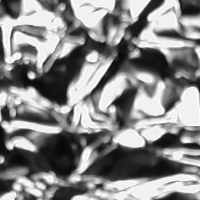}
    \includegraphics[width = 0.19\textwidth]{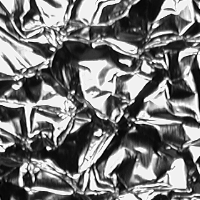}
    \includegraphics[width = 0.19\textwidth]{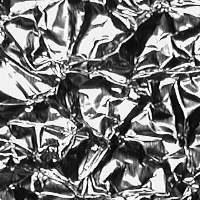}
    \includegraphics[width = 0.19\textwidth]{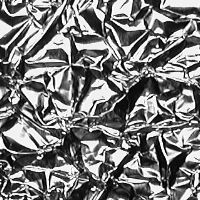}
    \includegraphics[width = 0.19\textwidth]{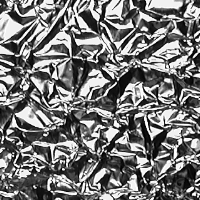}
    \includegraphics[width = 0.19\textwidth]{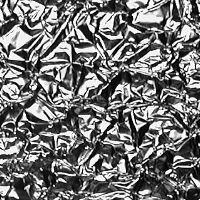}
    \includegraphics[width = 0.19\textwidth]{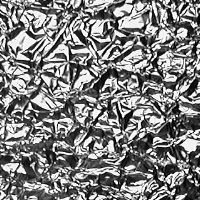}
    \includegraphics[width = 0.19\textwidth]{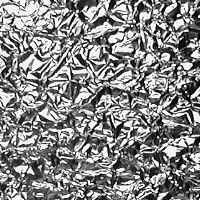}
    \includegraphics[width = 0.19\textwidth]{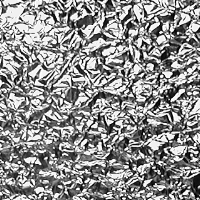}
    \caption{Scale variation in the KTH-tips dataset for the aluminium foil sample.}
    \label{fig:kth-scales}
\end{figure}

KTH-tips~\cite{fritz2004kth, hayman2004significance} consists of images of ten classes of textures under varying illumination, orientation, and scale. 
Texture classes are real materials (sandpaper,
crumpled aluminium foil, styrofoam, sponge), fabrics like corduroy, linen, and cotton, and natural structures like brown bread, orange peel, and cracker (Figure~\ref{fig:kth-classes}).
Each class has 81 images split in 9 scales. We use 40 images 
for training and the rest for testing.
In this dataset, textures fill the image window. Moreover, textures are spatially homogeneous by nature. 
Thus, no bounding box is needed.
A particularly interesting aspect of this dataset is scale variation within every class due to the varying viewing conditions, e.g. due to the changing distance between the camera and the object (Figure~\ref{fig:kth-scales}). 

As a baseline, we use Bruna's scattering network~\cite{bruna2012-2}, implemented using Kymatio~\cite{andreux2020kymatio}, see Appendix~\ref{appendix:scattering}. Following~\cite{andreux2020kymatio}, we use a depth of 2. We use the discrete rotation group with $M=4$ as for the Riesz feature representation. That is, filters are steered for the angles $\{0,\frac{\pi}{4},\frac{\pi}{2}, \frac{3\pi}{4}\}$. The maximal number of scales $J=7$ (see Appendix~\ref{appendix:scattering}) is selected to achieve global translation invariance. 
The scattering network outputs $365$ features.
For both representations, a PCA classifier (Appendix~\ref{appendix:PCA}) with the first 20 principal components is used.

\subsubsection{Experimental analysis of nonexpansiveness (Experiment 0)}
\label{sec:nonexpand:kth}
In Mallat's work \cite{mallat2012}, nonexpansiveness together with the energy decay of wavelets is the key to proving Lipschitz continuity to small deformations (Appendix~\ref{appendix:scattering}). 
Generally, missing nonexpansiveness was identified as the main reason why neural networks are sensitive to small deformations. However, in many applications, the output of the feature representations is scaled prior to training the classifier (e.g. SVM), or batch normalization is used. Feature scaling and batch normalization distort the nonexpansiveness. Hence, even if a scattering network is used, one should be careful in classifier design to preserve the nonexpansiveness of the scattering network.

Experimentally, we investigate how sensitive the performance of the Riesz feature representation is to changes in the nonexpansiveness constant $C$. For $M=4$, $C$ should be $\frac{1}{4}$ to achieve nonexpansiveness. For $C < \frac{1}{4}$, one has strict inequality for $S$, i.e. $||S(f)-S(g)|| < ||f-g||$, while for $C >\frac{1}{4}$, there exists a constant $C_1 = 4C > 1$ s.t. $||S(f)-S(g)|| \leq C_1||f-g||$. 
Hence, for every $C$, the Riesz feature representation $\Phi(f-g)$ remains bounded by $C_1||f-g||$ for some constant $C_1 > 0$. This is in contrast to feature scaling and batch normalization.

We found that the choice of $C$ did not have a large effect on the classification accuracy on the KTH-tips dataset (Table~\ref{tab:factorC}). Hence, in all experiments in the paper, we keep $C$ fixed to $1$. 


  \begin{table}[]
    \centering
    \begin{tabular}{|c|c|c|c|c|c|c|}
    \hline
     C & 0.125 & 0.25 & 0.5 & 1 & 2 & 4  \\
     \hline
      ours-PCA & 96.34 & 97.07  & 96.83 & 97.32 & 96.83 & 96.10 \\
     \hline
    \end{tabular}
    \caption{\textbf{(Experiment 0):} Accuracy of our Riesz representation (in $\%$) for scaling factor $C$ on the randomly split KTH-tips dataset. 
    }
    \label{tab:factorC}
\end{table}

\begin{table}[h]
    \centering
    \begin{tabular}{|c|c|c|c|c|c||c|}
    \hline
     & seed 42 & seed 21 & seed 10 & seed 5 & seed 0 & mean (std) \\
     \hline
     ours-PCA & 97.32 & 94.39 & 94.88 & 95.12 & 96.34 & 95.61 (1.196)
 \\
     Scattering-PCA & 97.56 & 95.61 & 98.05 & 96.10 & 98.05 & 97.06 (1.133)
 \\
     \hline
    \end{tabular}
    \caption{\textbf{(Experiment 1):} Random splitting for 5 seeds: accuracy (in $\%$). Our Riesz representation has depth 3 and uses 4 angles, resulting in 85 features. The scattering network has depth 2 and uses 4 angles (in total 365 features). }
    \label{tab:random-mix}
\end{table}

\subsubsection{Random splitting of the dataset (Experiment 1)}
This experiment tests the general performance of the methods on this dataset, i.e. how expressive features of two methods are in characterizing the textures.
Following the commonly applied approach, we split the dataset randomly into training and test sets without any regard for orientation, scale, or illumination.  
We report the accuracies for 5 random splittings in Table~\ref{tab:random-mix}.
The scattering network achieves slightly better results in this setting, but uses 4 times more features. However, random splitting of the dataset does not shed light on how useful or robust the representation is under completely unseen conditions, e.g. scales. 

\begin{table}[]
    \centering
    \begin{tabular}{|c|c|c|c|c|c|c|}
    \hline
     scaling factor  & 1.05 & 1.1 & 1.15 & 1.25 & 1.5 & 2 \\
     \hline
     ours-PCA   & 96.34 &  96.59 & 96.34 & 96.34  &  96.10 & 96.10 \\
     ours-PCA (cropping) & 95.37 &  94.63 & 95.37 &  95.37 & 93.41  & 86.10 \\
     scattering-PCA  & 97.56  & 96.34  & 95.37  & 94.15 & 81.95  &  65.37\\
     \hline
         \hline
     scaling factor & 0.5 & 0.75 & 0.85 & 0.9 & 0.95 & 1 \\
     \hline
     ours-PCA & 82.44 & 95.37 & 95.85 & 97.07 & 96.58 & 97.32 \\
     ours-PCA (padding) &  67.56 & 89.02 &  94.15 & 94.88 &  94.39 &  97.32 \\
     scattering-PCA & 65.12 & 90.00 & 96.34 & 96.10 & 96.83 & 97.56 \\
     \hline
    \end{tabular}
    \caption{\textbf{(Experiment 2):} Scale robustness for random splitting of the dataset: controlled scaling of test set. Accuracy in $\%$. The Riesz representation (ours) has depth 3 and uses 4 angles (in total 85 features). The scattering network has depth 2 and uses 4 angles (in total 365 features). 
    }
    \label{tab:upscale}
\end{table}


\begin{table}[]
    \vspace{0.1cm}
    \centering
    \begin{tabular}{|c|c|c|c|}
    \hline
     & train-first 40 & train-mid 40 & train-last 40 \\
     \hline
     ours-PCA & 77.07 & 86.82 & 88.75 \\
     scattering-PCA & 55.61 & 79.02 & 62.25 \\
     \hline
    \end{tabular}
    \caption{\textbf{(Experiment 3):} Scale robustness for scale dependent splitting of the dataset: accuracy (in $\%$). The Riesz representation (ours) has depth 3 and uses 4 angles (in total 85 features). The scattering network has depth 2 and uses 4 angles (in total 365 features)}
    \label{tab:scale-vary-train}
\end{table}

\subsubsection{Scale robustness for random splitting of the dataset (Experiment 2)}

The goal of this experiment is to test robustness of our representation with respect to 
rescaling of the image. For this purpose, images are scaled by factors from the range $[0.5,2]$ using spline interpolation. 
Scaling imitates small camera movements. 

Note that rescaling can destroy valuable texture information due to blurring or interpolation. Hence,  performance is expected to worsen as the rescaling factor deviates further from 1. 
For this experiment, the Riesz feature representation does not require any adjustments since it works on arbitrary image sizes.
In contrast, the scattering network requires either cropping (for upscaling) or reflective padding (for downscaling) to the fixed image size of $200 \times 200$.
To analyze the effect of cropping and padding on the performance of the method, we report the results for the Riesz feature representation both on the same input images as for the scattering network (with cropping/padding) and on the unchanged rescaled images (without cropping/padding).

For fixed seed, accuracy is shown in Table~\ref{tab:upscale} where scale 1 refers to the original image without rescaling. 
Here, we can notice that the Riesz feature representation without padding/cropping is significantly more stable to the larger scale variations than the scattering network.
Interestingly, results for the Riesz feature representation with cropping (Table~\ref{tab:upscale} top) are significantly better than for the scattering network, while in the case of padding (Table~\ref{tab:upscale} bottom) results are comparable to those of the scattering network. This implies that the padding disrupts the textural composition.

\subsubsection{Scale robustness for scale dependent splitting of the dataset (Experiment 3)}

Since scale information is known for this dataset, we can choose different scales for training and test sets and test the scale generalization ability this way. 
We test the performance of the methods on scales that are outside of the training set scale distribution. This is useful in settings where we are not able to collect all possible scales in the training set. This experiment tests larger scale variations than the previous one. Note that this type of scale dependent splitting of the dataset was previously done in \cite{lindeberg2020} for texture classification by matching scale equivariant image descriptors on the KTH-Tips2b dataset. 

Here, we conduct 3 experiments based on the selection of scales in the training set. The 40 training images are selected from either the smallest, the medium-sized or the largest scales in the dataset.

Scale variation for the aluminium foil class is shown in Figure~\ref{fig:kth-scales}. 
Results on the remainder of the dataset 
for each training set splitting are shown in Table~\ref{tab:scale-vary-train}.
The just described setting turns out to be the hardest due to completely unseen scales in the testing set. The Riesz representation reaches accuracies in the range $[0.77,0.88]$ which is an improvement over the scattering network with accuracies in the range of $8-25\%$ depending on the scenario.

\subsection{Experiments on CIFAR-10: importance of scale equivariance in deep networks} 

In this section, we analyze the scale equivariance property of our method and a deep learning network on CIFAR-10 \cite{krizhevsky2010cifar}, a dataset based on photographs of objects and animals with varying positions and backgrounds. This dataset consists of 60,000 images of size $32\times 32$ divided into 10 classes: airplane, automobile, bird, cat, deer, dog, frog, horse, ship, and truck. 
As deep network, we use Wide Residual Network (WRN) \cite{zagoruyko2016wide}, as it has achieved state-of-the-art results on this dataset. We used the implementation\footnote{\url{https://github.com/edouardoyallon/scalingscattering/tree/master/CIFAR/models}} of WRN from \cite{oyallon2017scaling}.

CIFAR-10 is split into 6 batches with 10,000 images, out of which one batch is reserved for testing. We used 5,000 images for validation and the rest for training. To ensure comparability, all methods are trained on the same dataset, without data augmentation\footnote{The reason for this is that the original manuscript does not specify details of the data augmentation. It only mentions that flipping and padding + cropping were done. The size of the dataset after augmentation is also not given. Hence, we restrict to the original data set.}.

We investigate the stability of the methods with respect to rescaling of the test set. A similar experiment on this dataset was first done in \cite{jansson22} for rescaling in the range $[0.5,2]$. However, we adopt a simpler approach: rescaling is done by relatively small factors $\{ 0.75, 0.875, 1.125, \\ 1.25\}$. 
When needed, the rescaled images are cropped (for upscaling) or padded (for downscaling) to the original size of $32\times 32$.
The goal of this experiment is to analyze the scale equivariance properties of the methods on natural images. Note that the selected scale variations are relatively small compared to what can happen in real-life scenarios or covered by \cite{jansson22}. Furthermore, the goal of this experiment is not to perform sole benchmarking on this dataset. 
We rather want to illustrate the lack of scale equivariance as a major weakness of well-established deep learning methods and suggest that these methods can highly benefit from incorporating equivariance properties.
The results are shown in Table \ref{tab:cifar-10}.

WRN was trained for 50 epochs using the Adam optimizer \cite{kingma14} with a learning rate of 0.01 which is halved every 7 epochs. The batch size is set to 50, and cross entropy loss function is used. The network has 10.96 million parameters. Table \ref{tab:cifar-10} shows us that WRN is not scale equivariant. We notice a large decrease in performance of around $39\%$ when using 0.75 as downscaling factor. On the other hand, the decrease in performance is less significant ($10\%$) when using 1.25 as upscaling factor. Nevertheless, this suggests that the state-of-the-art method WRN would not be able to generalize to unseen scales outside of the scale of the training set, despite of using almost 11 million parameters.
The authors in \cite{jansson22} offer a solution how scale equivariance properties of CNNs can be improved by introducing scale channels. In the scale interval $[0.75,1.25]$ accuracies for FovMax are reported to be around $85\%$, $90\%,$ and $75\%$ for scales $0.75$, $1$ and $1.25$, respectively, which is significantly more stable to scale variations than WRN.

\begin{table}[]
    \centering
    \begin{tabular}{|c||c|c||c||c|c||c|}
        \hline
         Method &  0.75 &  0.875 & 1 & 1.125  &  1.25 & \#param \\
         \hline
         WRN \cite{zagoruyko2016wide}   & 49.0 & 74.0 & 88.3 & 83.3 & 78.0 & 10.96M \\
         ours-SVM & 62.6 &  63.5  & 64.1 & 63.6 & 63.7 & $\sim$ \\
         ours-WRN & 69.3 & 70.3 & 72.1 & 71.0 & 71.0 & 10.80M \\
         \hline
    \end{tabular}
    \caption{Classification accuracies (in $\%$) on CIFAR-10 dataset trained without any data augmentation to enable comparability. Scale 1 refers to the original test set from CIFAR-10 dataset.}
    \label{tab:cifar-10}
\end{table}

For the Riesz feature representation, we use the same setup as in Section \ref{mnist:exp} up to the final pooling layer. The authors of \cite{sifre2013, oyallon2017scaling} prove the benefit of more localized features in comparison to a single global one at the expense of losing the translation invariance. Hence, we follow the same approach: we downscale every feature map of size $32\times 32$ to $8\times 8$. Every downscaled feature map is vectorized, concatenated in a single vector, and used as an input for the SVM classifier. The results in Table \ref{tab:cifar-10} show that WRN outperforms our method on the standard CIFAR-10 test set. This is the consequence of having way more degrees of freedom (trainable parameters) and is in agreement with the literature \cite{oyallon2017scaling}. However, our method is significantly more stable to variations in the scale showing a minimal decrease in the performance ($0.5-1.5 \%$) compared to WRN ($5-39\%$).

In summary, WRN achieves better performance than our method on the original CIFAR-10 test set, while showing significant instabilities to the scale variations due to the lack of scale equivariance. 
One way to combine the best from both worlds is through a hybrid approach \cite{oyallon2017scaling}: build a deep WRN on the top of the scale equivariant Riesz feature representation that was downscaled as described above. We use the same training setup as for the WRN from above. Table \ref{tab:cifar-10} reports the first results for this scenario. Here, stability to the variations in scale is significantly improved compared to WRN ($1-3 \%$ vs $5-39\%$), while overall classification accuracy is increased by roughly $8\%$ compared to the SVM classifier on the Riesz feature representation. Hence, the hybrid approach benefits from both approaches.   

\section{Discussion}

The Riesz feature representation is a hand-crafted feature extractor based on stacking quadrature filters using first and second order Riesz transforms in a hierarchy. 
This representation creates features that are scale and translation equivariant 
and avoids sampling or discretization of the scale dimension. 

As a result, the number of features is significantly reduced compared to other representations such as scattering networks. Our representation contains only 85 features, which is a reduction by factor 4 compared to the corresponding scattering representation.

A consequence of scale equivariance is the generalization to unseen scales. Hence, this work complements and reflects on the previous research efforts on scale generalization \cite{barisin22riesz, lindeberg2020,lindeberg2022scale, jansson22}. We show this empirically on the MNIST Large Scale dataset by training an SVM classifier on the Riesz feature representation on a fixed scale, and testing on unseen scales that differ significantly from the one in the training set. 
Moreover, our representation proves to be useful for texture classification on the KTH-tips dataset with a small training set. When including all scales in the training set, the Riesz feature representation yields results comparable to those of the scattering network. However, when splitting training and testing set along scales, the Riesz feature representation turns out to be more robust. 

Important benefits of scattering networks 
are that they avoid the time-consuming training process and parameter tuning as needed for CNNs and work well even for small training sets. Only 40 and 500 training images per class were needed for KTH-tips and MNIST Large Scale, respectively, to achieve accuracies over 0.9. 
The most important benefit of using the Riesz transform is however that robustness to scale variations is ensured without additional training. This was proven to be true even for scale variations in the natural images of the CIFAR-10 dataset.
However, the experiments on CIFAR-10 show that deep networks still have better general discriminative properties than the proposed approach on the original test. Hence, future work should aim to close this gap: to investigate whether it is possible to improve the discriminative power of this representation while preserving the majority of the mathematical properties.

All types of scattering networks including our approach rely on using only the amplitude to design feature representations, i.e. the phase is completely discarded. In one of the first works on scattering networks~\cite{bruna2012-2} it was argued that phase can be reconstructed from amplitude information by solving the so-called phase recovery problem~\cite{waldspurger2012}. However, phase recovery is a non-convex optimization problem. Hence, the question remains open, how to use 
the phase information in the framework of scattering networks and whether that would result in improved capabilities of this class of methods.

The main challenge in applying the
Riesz feature representation to a wider range of problems (e.g. object detection) is to devise a scale equivariant bounding box algorithm. This is required for images with complex scenes which contain several objects belonging to different classes. A scale equivariant bounding box algorithm is hence subject of future work. 

Finally, naturally, our representation can be used to create hybrid trainable representations by combining them with building blocks of deep neural networks, see e.g.~\cite{cotter2019learnable,oyallon2017scaling,oyallon2018scattering}. The first results on the CIFAR-10 dataset are promising: improved classification accuracy compared to the standard SVM built on the Riesz feature representation and more stable performance with respect to scale variations compared to the standard deep network.

\appendix



\section{Visualizations of Riesz representation feature maps}
\label{appendix:visualization}

\begin{figure}[!h]
    \centering
         \begin{subfigure}{0.24\textwidth}
    \includegraphics[width=\textwidth]{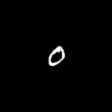}
    \caption{Input image}
    \end{subfigure}
    \\
         \begin{subfigure}{0.24\textwidth}
    \includegraphics[width=\textwidth]{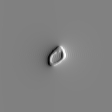}
     \caption{$\mathcal{H}_0\textcolor{white}{\psi_{\frac \pi 4}}$}
    \end{subfigure}
         \begin{subfigure}{0.24\textwidth}
    \includegraphics[width=\textwidth]{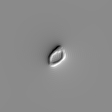}
    \caption{$\mathcal{H}_{\frac{\pi}{4}}$}
    \end{subfigure}
         \begin{subfigure}{0.24\textwidth}
    \includegraphics[width=\textwidth]{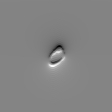}
    \caption{$\mathcal{H}_{\frac{\pi}{2}}$ }
    \end{subfigure}
         \begin{subfigure}{0.24\textwidth}
    \includegraphics[width=\textwidth]{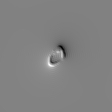}
    \caption{$\mathcal{H}_{\frac{3\pi}{4}}$}
    \end{subfigure}
    \\
         \begin{subfigure}{0.24\textwidth}
    \includegraphics[width=\textwidth]{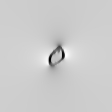}
    \caption{$\mathcal{H}_{0}^{(2)}\textcolor{white}{\psi_{\frac \pi 4}}$}
    \end{subfigure}
         \begin{subfigure}{0.24\textwidth}
    \includegraphics[width=\textwidth]{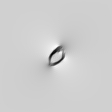}
    \caption{$\mathcal{H}_{\frac{\pi}{4}}^{(2)}$}
    \end{subfigure}
         \begin{subfigure}{0.24\textwidth}
    \includegraphics[width=\textwidth]{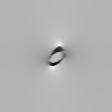}
    \caption{$\mathcal{H}_{\frac{\pi}{2}}^{(2)}$}
    \end{subfigure}
         \begin{subfigure}{0.24\textwidth}
    \includegraphics[width=\textwidth]{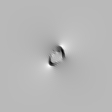}
    \caption{$\mathcal{H}_{\frac{3\pi}{4}}^{(2)}$}
    \end{subfigure}
    \caption{Steered first and second order Riesz transforms applied to a digit from the MNIST Large scale dataset. 
     White represents a high positive filter response, gray is close to 0, and black is a negative filter response. Here, we slightly abuse the notation by writing $\mathcal{H}_{\phi }$ rather than $\mathcal{H}_{v }$ for $v=(\cos \phi, \sin \phi)$. See Section~\ref{section:rt:steer} for details. 
     }
    \label{fig:riesz:transform:mnist}
\end{figure}

\begin{figure}[!h]
    \vspace{0.1cm}
    \centering
    \begin{subfigure}{0.24\textwidth}
    \includegraphics[width=\textwidth]{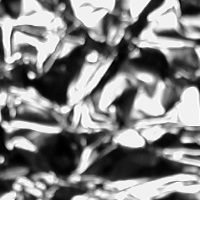}
    \caption{Input image}
    \end{subfigure}
    \\
        \begin{subfigure}{0.24\textwidth}
    \includegraphics[width=\textwidth]{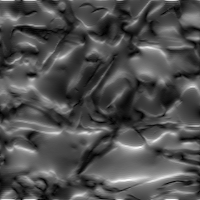}
    \caption{$|f\ast\psi_0| \textcolor{white}{\psi_{\frac \pi 4}}$}
        \end{subfigure}
        \begin{subfigure}{0.24\textwidth}
    \includegraphics[width=\textwidth]{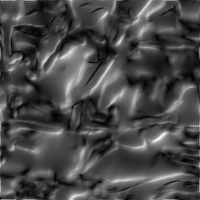}
    \caption{$|f\ast\psi_{\frac \pi 4}|$}
        \end{subfigure}
        \begin{subfigure}{0.24\textwidth}
    \includegraphics[width=\textwidth]{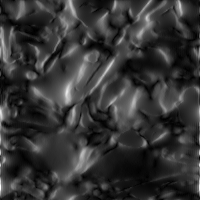}
        \caption{$|f\ast\psi_{\frac \pi 2}|$}
        \end{subfigure}
        \begin{subfigure}{0.24\textwidth}
    \includegraphics[width=\textwidth]{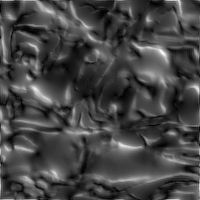}
        \caption{$|f\ast\psi_{\frac{3\pi}{ 4}}|$}
        \end{subfigure}
    \\
        \begin{subfigure}{0.24\textwidth}
    \includegraphics[width=\textwidth]{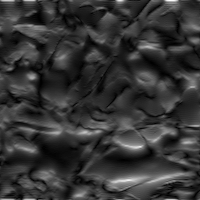}
    \caption{$||f\ast\psi_0|\ast\psi_0|\textcolor{white}{\psi_{\frac \pi 4}}$}
        \end{subfigure}
        \begin{subfigure}{0.24\textwidth}
   \includegraphics[width=\textwidth]{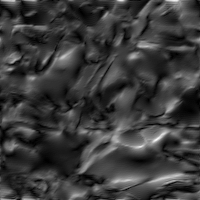}
   \caption{  $||f\ast\psi_0|\ast\psi_{\frac{\pi}{4}}|$}
       \end{subfigure}
       \begin{subfigure}{0.24\textwidth}
   \includegraphics[width=\textwidth]{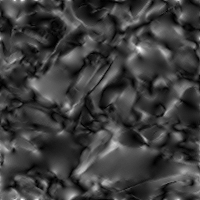}
   \caption{$||f\ast\psi_0|\ast\psi_{\frac{\pi}{2}}|$}
       \end{subfigure}
       \begin{subfigure}{0.24\textwidth}
    \includegraphics[width=\textwidth]{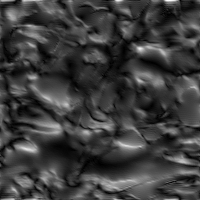}
    \caption{ $||f\ast\psi_0|\ast\psi_{\frac{3\pi}{4}}|$}
        \end{subfigure}
    \\
        \begin{subfigure}{0.24\textwidth}
    \includegraphics[width=\textwidth]{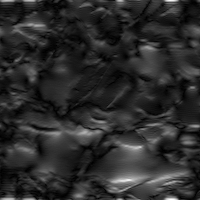}
    \caption{\tiny $|||f\ast\psi_0|\ast\psi_0|\ast\psi_0|\textcolor{white}{\psi_{\frac \pi 4}}$}
        \end{subfigure}
        \begin{subfigure}{0.24\textwidth}
   \includegraphics[width=\textwidth]{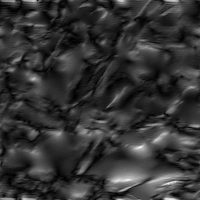}
   \caption{\tiny$|||f\ast\psi_0|\ast\psi_0|\ast\psi_{\frac{\pi}{4}}|$}
              \end{subfigure}
       \begin{subfigure}{0.24\textwidth}

   \includegraphics[width=\textwidth]{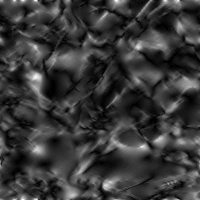}
   \caption{\tiny$|||f\ast\psi_0|\ast\psi_0|\ast\psi_{\frac{\pi}{2}}|$}
              \end{subfigure}
       \begin{subfigure}{0.24\textwidth}
    \includegraphics[width=\textwidth]{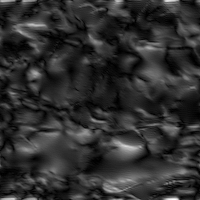}
    \caption{ \tiny $|||f\ast\psi_0|\ast\psi_0|\ast\psi_{\frac{3\pi}{4}}|$}
               \end{subfigure}
    \caption{Riesz feature representation prior to pooling on the aluminum texture from the KTH-tips dataset. Top:  input image $f$. Second to fourth row: Riesz representations at depth 1, 2 and 3, respectively.
    White corresponds to a high positive filter response, while black denotes a filter response close to~0. 
    }
    \label{riesz:representation:kth}
\end{figure}

Here, features extracted by the steered base functions derived from the Riesz transform are visualized. 
Figure~\ref{fig:riesz:transform:mnist} features an image of a digit from the MNIST Large dataset.
Figure~\ref{riesz:representation:kth} shows selected feature maps from the Riesz feature representation up to depth $3$  and prior to pooling for the aluminum foil from the KTH-tips dataset.

\section{Introduction to scattering networks}
\label{appendix:scattering}
Scattering networks are based on applying the wavelet transform and modulus non-linearity in a hierarchy creating a deep feature representation. 
Features from every level of the hierarchy are used to construct a group invariant representation by a local pooling operation on the group domain.
Here, we give an overview of the locally translation invariant scattering network from~\cite{bruna2012-2}.

First, the Morlet mother wavelet $\psi_M(x):\R^2\to\C$ is defined as an oriented bandpass complex function
$$\psi_M(x) = \alpha (e^{i \langle x, \xi \rangle}-\beta)e^{-|x|^2/(2\sigma^2)}.$$
The parameter $\xi \in \R^2, ||\xi||_2=1$ 
defines the orientation of the wavelet, while $\sigma$ controls its scale. Furthermore, $\beta<<1$ is chosen such that $\int \psi_M(x) dx = 0$. 
The mother wavelet is rotated by $r \in SO(2)$ and scaled (dilated) by $2^j, j \in \Z$, to create diverse features:
$$ \psi_\lambda(x)=2^{-2j}\psi_M(2^{-j}r^{-1}x),$$
for $\lambda=2^jr$. 

From now on, let $\mathcal{P}=\{ 2^{j}r \quad | \quad j \in \{1, \cdots, J \}, r \in G_M\}$ denote the space of all selected scale and rotation transformations\footnote{Here, $J$ denotes the total number of scales to be considered, while $G_M=\{k\frac{\pi}{M}, k\in\{0, 1,\cdots,M-1\}\}$ is a finite rotation group  where $M\in \N$ controls the discretization of the group. Generally, $\mathcal{P}$ is a discrete version of the more general space of scale and rotation transformations $\mathcal{P}^*=\{ 2^{j}r \quad | \quad j \in \Z, r \in SO(2)\}$.} of the wavelet $\psi_M$. Next, non-linearity is needed to construct a (non-trivial) translation invariant representation\footnote{For image $f$, it follows $\int_{\R^2} (f \ast \psi_M)(x)dx = \int_{\R^2} f(y) \big(\int_{\R^2} \psi_M(x-y) dx \big) dy= 0 $. Hence, for every $f$ one gets the trivial translation invariance.} due to $\int \psi_M(x) dx = 0$, see~\cite{bruna2012-2} for details. 
We select the modulus of the complex number
$$A(y_r+iy_i)=\sqrt{y_r^2+y_i^2},$$
as it is a pointwise nonexpansive non-linearity operator which preserves signal energy.
For an image $f\in L_2(\R^2)$, features at depth $1$ for rotation and scale $\lambda=2^jr \in \mathcal{P}$ are extracted using the \textbf{scattering transform} operator $U_\lambda:L_2(\R^2) \to L_2(\R^2)$
$$ U_\lambda f = A(f \ast \psi_\lambda).$$ 

The output of this scattering transform is called a \textbf{scattering coefficient}. To create a hierarchy of scattering coefficients, scattering transforms are applied to the scattering coefficients from the previous step. We refer to these as higher order scattering coefficients. Generally, for depth $m$ we apply the scattering transform $m$ times sequentially. This operator is defined by a sequence $p=(\lambda_1,\cdots, \lambda_m) \in \mathcal{P}^m$ of $m$ scales and rotations via
$$ U_p f = U_{\lambda_m} \cdots U_{\lambda_1}f.$$

Operators $U_p$ for $p \in $ transform input image into the feature maps which have the same size as the input image. Hence, similarly as in Section \ref{sec:pooling} pooling operations are needed to achieve translation invariance. Based on the choice of the pooling operator one can achieve either global or local translation invariance \cite{bruna2012-2}.


To calculate (global) translation invariant scattering coefficients, a global pooling operator $\overline{S}_p$ can be applied 
$$ \overline{S}_p(f) = \int U_pf(x) dx.$$

In some applications, it is beneficial to have features invariant with respect to translations by up to $2^j$ pixels for $j\in \N$, where an optimal $j$ can be determined through cross-validation. This local translation invariance can be achieved by convolution with a Gaussian kernel $g_\sigma$ with $\sigma=2^j$ which yields
$$ S_p(f)(x) = ( U_p f\ast g_\sigma)(x).$$
According to scale space theory, the convolution with $g_{\sigma}$ removes all information on scales smaller than $\sigma=2^j$. Hence, 
the representation can be downsampled by $2^j$ as local translation invariance up to $2^j$ pixels is guaranteed. 

Finally, scattering coefficients can be written as $W_0(f)=f\ast g_{2^j}$, $W_1(f)=\big( S_\lambda(f)\big)_{\lambda \in \mathcal{P}}$, and $W_k(f)=\big( S_p(f)\big)_{p \in \mathcal{P}^k}$. 
Hence, the locally translation invariant scattering representation can be written compactly as 
$$\Phi_K(f) = \Big( W_k(f) \quad | \quad k=0, \cdots, K  \Big).$$

The mapping $\Phi_K$ is called a \textbf{scattering network}. Scattering networks are claimed to be a proxy for convolutional neural networks as they use cascades of the same building blocks: convolutional filters, non-linearities, and pooling operators.
The main difference is that convolutional filters in scattering networks are deterministic, i.e. no training procedure is required.

Next, we present two important properties related to signal properties which characterize a scattering network.

\subsection*{Energy preservation and Lipschitz continuity to small deformations:} 

In~\cite{bruna2012-2} it was shown that scattering networks
induce energy preservation in the sense that 
$$ ||f||^2 = \sum_{k=0}^\infty \sum_{p,|p|=k} ||S_p(f)||^2.$$
Let $\tau : \R^2 \to \R^2$ be a small deformation displacement field, i.e.,  $||\nabla \tau||_{\infty} \leq 1$. 
Then for $\Phi_\infty =  \cup_{k=0}^\infty W_k(f)$, $L_\tau(f)(x)= f(x+\tau(x))$ and under the assumption that $g_{2^j}, \psi_M$ 
and their first two derivatives have a decay of $O\Big((1+|x|)^{-(d+2)}\Big)$, Lipschitz continuity to small deformations $\tau$ holds. That is
\begin{equation}
\label{lipschitz:cont:eq}
    || \Phi\big(L_\tau(f)\big) - \Phi(f)|| \leq 
 C||f|| \text{ }||\nabla \tau||_\infty.
\end{equation} 
These claims were proven by Mallat in~\cite{mallat2012}.

\renewcommand{\thesubsubsection}{\arabic{subsection})}
\subsubsection*{Lipschitz continuity to small deformations for the Riesz feature representation}
\label{appendix:lipschitz}
The Riesz transform does not satisfy Mallat's decay assumption~\cite{mallat2012}. Instead, we derive Lipschitz continuity from non-expansiveness using a simple trick. In fact, it is enough to apply a smoothing operator which regularizes the representation. Formally, let $\gamma:L^2(\R^d) \to L^2(\R^d)$ be a smoothing operator that is Lipschitz continuous to small deformations, i.e. it satisfies equation \eqref{lipschitz:cont:eq}. 
Let $\Phi$ be an arbitrary representation that is nonexpansive. Then $\Phi \circ \gamma$ is Lipschitz continuous to small deformations: 
$$ ||\Phi (\gamma(f)) - \Phi \big(\gamma( L_{\tau}(f))\big)|| \leq || \gamma (f) - \gamma \big(L_{\tau}(f)\big) || \leq C ||f|| \text{ } ||\nabla \tau ||_{\infty} .$$
Hence, a smoothing operator like a convolution with a Gaussian kernel is needed to regularize the input. 
The smoothing factor $\sigma$ should however not be too large in order to not destroy important structural information.

\section{Multilayer perceptron as classifier}
\label{appendix:mlp}
A multilayer perceptron (MLP) is a fully connected feedforward neural network whose layers consist of a fully connected layer (or 1d convolution), batch normalization, and non-linear activation (ReLU).
We explore the MLP as an alternative to the SVM 
and
feed it
with the output of the Riesz representation. 

\textbf{Details on training:} We apply the following hyperparameters: cross entropy loss, batch size $50$, Adam optimizer~\cite{kingma14} with initial learning rate $0.001$. Number of epochs and learning rate decay are adapted to the size of the training set.
For a training set of $1,000$ images, we use $75$ epochs and step decay for which the learning rate halves every $10$ epochs.
For a training set of $50,000$ images, we use $30$ epochs and halve the learning rate every $4$ epochs.

\textbf{Ablation study on Riesz feature representation with MLP classifier:}
We vary the fineness of discretization of the rotation group and the depth of the Riesz feature representation to understand how to optimally balance Riesz feature representation size and accuracy.
An MLP with 2 hidden layers each having $128$ channels is used as a classifier for every parameter configuration of the Riesz feature representation. 
We restrict to the scales $\{ 0.5,1,2,4,8\}$ to reduce runtime. Results are summarized in Table~\ref{tab:ablationRieszRep},
details on architecture and number of parameters in Table~\ref{tab:ablation:details}. 

In the chosen architecture, the output of the Riesz feature representation is the first channel. Therefore, the number of parameters of the MLP depends on the parameter configuration of the Riesz feature representation. 
We use again the Riesz feature representation of depth 3 and with 4-angle rotation group as baseline. Results are reported using the full training set of $50,000$ images.

The increase in depth from 3 to 4 improves the results by not more than $1\%$, while the number of parameters increases by a factor of 4. This indicates that the new features are non-informative and redundant.
Similar effects can be observed when enlarging the finite rotation group from 4 to 8 angles. Simultaneously reducing the depth to 2 yields results slightly worse (around $2\%$) than those of the baseline while the number of features stays roughly the same: $85$ vs $73$. These results indicate that for this problem between $70$ and $100$ features should be optimal.

\begin{table}[h]
    \centering
    \begin{tabular}{|r|r|r|c|c|c|c|c|}
     \hline
     \multicolumn{3}{|c|}{MLP/scale} & 0.5 & 1 (t) & 2 & 4 & 8  \\
     \hline
     \# training images & depth & \# angles & & & & &\\
     \hline
     \hline
     1,000 & 3 & 4 & 70.72 & 89.00 & 87.48 & 87.30 & 59.79  \\ 
     \hline
     1,000 & 4 & 4 & 68.40 & 90.85 & 90.39 & 90.29 & 49.60  \\ 
     \hline
     \hline
     50,000 & 2 & 8 & 79.45 & 94.08 & 93.64 & 93.64 & 67.08  \\ 
     \hline
     50,000 & 3 & 4 & 82.78 & 96.23 & 95.60 & 95.61 & 67.92  \\ 
     \hline
     50,000 & 3 & 8 & 83.64 & 96.77 &  96.42 & 96.48 & 73.77  \\ 
     \hline
     50,000 & 4 & 4 & 80.23 & 96.83 & 96.48 & 96.50 & 64.16  \\ 
     \hline
\end{tabular}
    \caption{Results (accuracies in \%) of the ablation study for the MLP on MNIST Large Scale. The test set consists of 10,000 images per scale. Trained on scale 1.}
    \label{tab:ablationRieszRep}
\end{table}

\begin{table}[h]
    \vspace{0.1cm}
    \centering
    \begin{tabular}{|c|c|c|c|c|}
    \hline
     & architecture & parameters \\
     \hline
     MLP depth 2 angle 8 & 73-128-128-10 & 27,786 \\
     \hline
     MLP depth 3 angle 4 & 85-128-128-10  & 29,322\\ 
     \hline
     MLP depth 3 angle 8 & 585-128-128-10 & 93,322 \\ 
     \hline
     MLP depth 4 angle 4 &  341-128-128-10 & 62,090   \\
     \hline
\end{tabular}
    \caption{Details of the applied MLPs with 2 hidden layers.
    Note that affine batch normalization was used for the two hidden layers in MLP, which increases the number of trainable parameters.
    }
    \label{tab:ablation:details}
\end{table}

\section{PCA classifier}
\label{appendix:PCA}
For the sake of self-containedness, we provide details of Bruna's PCA classification~\cite{bruna2013}.
Let $n \in \N$ be the number of training images per class, $C\in\N$ the number of classes, and let $\{I_1, \cdots, I_n\}$ denote the images belonging to class $c$. Let $E_c(\Phi)$ 
denote the expected value of the $P$-dimensional Riesz feature representation for an image from class $c$. Practically, it is estimated by the mean of $\{\Phi(I_1), \cdots, \Phi(I_n)\}$.
The goal of the PCA is to approximate the centered data $\Phi - E_c(\Phi)$ on class $c$ by its projection to a lower  ($d$-) dimensional space $V_c$ for $d << P$.  The subspace $V_c$ is then said to be spanned by $d$ principal components.
After computing $V_c$ for every class $c$, the projection errors when projecting a given input image $I$ on each $V_c$ can be calculated. 
The smaller the projection error, the more likely is the image to belong to that class. Hence, a classification rule can be easily derived:
Let $P_{V_c}$ be the projection operator for class $c$. The define the PCA classifier $k_{PCA}$ for image $I$ by
$$k_{PCA}(I) = \argmin_{c}|| \Phi(I) - E_c(\Phi) - P_{V_c}(\Phi(I)- E_c(\Phi))||_2. $$
We used the PCA implementation \textit{sklearn}~\cite{scikit-learn} in Python.

\section{Technical supplements of Section~\ref{base-function5:1} }
\label{appendix:technical}
\textbf{Non-expansiveness of $\psi$:}

\begin{lemma}
\label{two:ineq:lemma}
The following statements hold for $f \in L_2(\R^d)$:
\begin{enumerate}
    \item $||\mathcal{H}_r (f)||^2 +  ||\mathcal{H}_{r+\pi/2} (f)||^2 \leq  ||f||^2 $
    \item $||\mathcal{H}_r (f)||^2 \leq ||f||^2,$
    \item $||\mathcal{H}^{(2)}_r(f)||^2 \leq ||f||^2.$
\end{enumerate}
\end{lemma}
\begin{proof}
    Here, we will use the following notation: let $r=(\cos \phi, \sin \phi)$, then $r+\pi/2:=(\cos(\phi+\frac{\pi}{2}), \sin(\phi+\frac{\pi}{2}))$.
    \begin{enumerate}
    \item
  \begin{align*}
    &||\mathcal{H}_r (f)||^2 +  ||\mathcal{H}_{r+\pi/2} (f)||^2 = \\
    &= ||\cos\phi \mathcal{R}_1(f) + \sin \phi \mathcal{R}_2(f)||^2 +  ||\cos(\phi+\pi/2) \mathcal{R}_1(f) + \sin(\phi+\pi/2) \mathcal{R}_2(f)||^2 \\
    &\leq ||\cos\phi \mathcal{R}_1(f)||^2 + ||\sin \phi \mathcal{R}_2(f)||^2 +  ||\cos(\phi+\pi/2) \mathcal{R}_1(f)||^2 + \\ 
    & \quad  + ||\sin(\phi+\pi/2) \mathcal{R}_2(f)||^2 = 
     (\cos^2(\phi) + \sin^2(\phi) ) \big(|| \mathcal{R}_1(f)||^2 + || \mathcal{R}_2(f)||^2 \big) = 
    \\
    &=|| \mathcal{R}_1(f)||^2 + || \mathcal{R}_2(f)||^2 \overset{\mathrm{\eqref{parseval:thm}}}{=} || f||^2.
    \end{align*}
    \item Follows immediately from 1.
\item This is proven using the inequality in 2. and the fact that $\mathcal{H}^{(2)}_r (f) = \mathcal{H}_r \big( \mathcal{H}_r (f) \big)$.    
    We even have
    \begin{align*}
        ||\mathcal{H}^{(2)}_r (f)||^2 +||\mathcal{H}^{(2)}_{r+\pi/2} (f)||^2 &= || \mathcal{H}_r \big( \mathcal{H}_r (f) \big)||^2 + || \mathcal{H}_{r+\pi/2} \big( \mathcal{H}_{r+\pi/2} (f) \big)||^2 
        \\
        &\leq || \mathcal{H}_r (f) \big)||^2 + || \mathcal{H}_{r+\pi/2} (f) \big)||^2  \leq ||f||^2.
    \end{align*}
       \end{enumerate}
\end{proof}
 \textbf{Zero integral of base function $\psi$:} 
Note that convolution with the Riesz kernel is not defined at 0. We therefore state the zero integral property using the Cauchy principal integral $\text{p.v.}$.
  \begin{lemma}
 \label{lemma:zero:int}
     $$p.v. \int_{\R^2} \psi(x) dx = 0.$$ 
 \end{lemma}
\begin{proof}
    From Equation (\ref{eq:base:function}) we have
  $\int_{\R^2} \psi(x) dx = i \cdot \int_{\R^2} r_1(x) dx ´+  \int_{\R^2} r^{(2,0)}(x) dx$. Thus, the claim can be proven for real and imaginary part, separately.
  The kernel $r_1$ in the real part is anti-symmetric, i.e. $r_1(x) = - r_1(-x)$ and hence $\text{p.v.}\int_{\R} r_1(x) dx = 0$.
  In the imaginary part, we have for the kernel $r^{(2,0)} = r_1 \ast r_1$ and hence
  \begin{align*}
      \text{p.v.}\int_{\R^2} r^{(2,0)}(x) dx &= \text{p.v.}\int_{\R^2} \text{p.v.}\int_{\R^2} r_1(t) r_1(x-t) dt dx = \\
      &=\text{p.v.}\int_{\R^2} r_1(t) \Big(\text{p.v.}\int_{\R^2} r_1(x-t)dx\Big) dt = \\
      &= [y=x-t \text{ change of variables}] =  \text{p.v.}\int_{\R^2} r_1(t)\Big( \int_{\R^2} r_1(y)dy \Big) dt = \\
      &=\text{[$r_1$ is anti-symmetric]} =0.
  \end{align*}
\end{proof}



\bibliographystyle{elsarticle-num}
\bibliography{references}

\end{document}